\documentclass{article}

\usepackage{microtype}
\usepackage{graphicx}
\usepackage{subcaption}
\usepackage{booktabs} 
\usepackage{placeins}

\usepackage{hyperref}

\usepackage[accepted]{icml2026}

\usepackage{amsmath}
\usepackage{amssymb}
\usepackage{mathtools}
\usepackage{amsthm}

\usepackage[capitalize,noabbrev]{cleveref}

\theoremstyle{plain}

\theoremstyle{definition}

\theoremstyle{remark}

\usepackage[textsize=tiny]{todonotes}

\icmltitlerunning{Viral Proteins Reveal Geometry of Protein Language Models}

\begin{document}

\twocolumn[
  \icmltitle{Viral Proteins Reveal Geometry of Protein Language Models}



  \icmlsetsymbol{equal}{*}

\begin{icmlauthorlist}
  \icmlauthor{Arthur Bigot}{eth,harvardccb}
  \icmlauthor{Harmon Bhasin}{twentytwo}
  \icmlauthor{Core Francisco Park}{harvard,prior}
  \icmlauthor{Eugene Shakhnovich}{harvardccb}
  \icmlauthor{Dianzhuo Wang}{harvardccb,twentytwo}
\end{icmlauthorlist}

\icmlaffiliation{eth}{Department of Computer Science, ETH Zurich, Zurich, Switzerland}
\icmlaffiliation{twentytwo}{TwentyTwo, USA}
\icmlaffiliation{harvard}{Harvard University, Cambridge, MA, USA}
\icmlaffiliation{prior}{Prior Computers, USA}
\icmlaffiliation{harvardccb}{Department of Chemistry and Chemical Biology, Harvard University, Cambridge, MA, USA}
\icmlcorrespondingauthor{Arthur Bigot}{bigot.arthur@gmail.com}
\icmlcorrespondingauthor{Eugene Shakhnovich}{shakhnovich@chemistry.harvard.edu}
\icmlcorrespondingauthor{Dianzhuo Wang}{john@twentytwo.bio}
  \icmlkeywords{Machine Learning, ICML, Protein Language Models, Biology, pLMs, Neural Networks, Deep Learning, Interpretability}

  \vskip 0.3in
]



\printAffiliationsAndNotice{}  

\begin{abstract}
Protein language models are trained on highly imbalanced datasets, raising the question of how they represent underrepresented biological sequences. Using viral proteins as a case study across ESM model families, we identify a dominant \emph{nativeness axis} in embedding space, aligned with masked-reconstruction perplexity, that orders sequences from well-modeled cellular proteins through viral proteins to shuffled and random sequences. Scaling contracts this axis unevenly across viral families. Despite this, protein language model embeddings retain viral-specific signal: viral proteins remain linearly separable beyond zero-shot perplexity and shallow sequence features. Together, these results suggest that pLM representations are structured by a general notion of nativeness while preserving information specific to distinct biological groups.
\end{abstract}

\section{Introduction}

Protein language models (pLMs) are powerful tools for learning general-purpose protein representations. Trained on large sequence databases, they are used for structure prediction \citep{lin2023esm2}, inverse folding \citep{hsu2022if}, and function prediction \citep{dallago2021flip, feldman2025biophysically, meier2021esm1v, notin2023proteingym}. Their strong performance has motivated a growing interest in understanding what these models learn, including recent work using mechanistic interpretability \citep{adams2025mechanistic, simon2025interplm, silberg2025functional}. However, much less is known about how pLMs represent biological groups that are functional but underrepresented and evolutionarily distinct from cellular proteins.

Viral proteins provide a useful case study for how pLMs represent underrepresented biological groups. Compared with proteins from cellular organisms, they are less abundant in standard pretraining data and are shaped by different evolutionary constraints, including host dependence, high mutation rates, compact genomes, and multifunctionality \citep{simmonds2019prisoners, Gurev2025.08.04.668549, tokuriki2009viral}. Benchmark results reflect this gap: pLMs struggle on viral mutation-effect prediction, despite strong performance on standard, mostly non-viral benchmarks \citep{Gurev2025.08.04.668549}.

In this work, we analyze viral protein representations across model scales to answer two questions: \emph{First, is viral separation mainly explained by lower nativeness, meaning that viral proteins are less well modeled by the distribution learned during pretraining? Second, beyond nativeness, do embeddings retain viral-specific information?} We answer the first question by analyzing the embedding geometry across model families through principal component analysis (PCA) and masked-reconstruction perplexity, which we use as a model-relative measure of nativeness. We answer the second question by comparing linear probes trained on mean-pooled embeddings with shallow sequence-based classification baselines and zero-shot classifier based on masked-reconstruction perplexity. Our main contributions are summarized as follows:

\begin{itemize}
    \item We identify a dominant nativeness axis in pLM representation space that closely tracks reconstruction difficulty and explains much of the viral to cellular protein shift.
    \item We show that scaling reduces viral displacement along this axis unevenly across human viral families, with some families moving toward the native region and others remaining persistently displaced.
    \item We show that embeddings retain viral signal beyond nativeness: linear probes outperform perplexity-only classifiers and shallow baselines across model families and scales.
\end{itemize}
\clearpage
\section{Related Work}



\paragraph{pLM scaling laws}
Language model loss and downstream performance improve predictably as model size, data, and compute are scaled \citep{kaplan2020scaling, hoffmann2022training}. In protein language models, scaling has similarly been associated with stronger biological representations and the emergence of capabilities useful for structure prediction, fitness prediction, function prediction, and protein generation \citep{rives2021biological, lin2023evolutionary, nijkamp2023progen2, hayes2025simulating,wangpnas,huot2025predicting}. However, it remains unclear how scaling changes the representation of underrepresented and evolutionarily distinct groups such as viral proteins.

\paragraph{Perplexity as a proxy for protein nativeness}
pLMs have been shown to encode biological constraints \citep{simmonds2019prisoners, Gurev2025.08.04.668549, tokuriki2009viral} beyond simple amino acid statistics, including structural contacts, family structure, and useful residue-level representations learned from sequence-only objectives \citep{rao2021transformer, rives2021biological}. As a result, pseudo-log-likelihood and pseudo-perplexity scores from masked pLMs have been widely used as zero-shot proxies for mutational effects, functional tolerance, viral escape, and generated-protein filtering \citep{salazar2020masked, meier2021language, hie2021learning, verkuil2022language, madani2023large, notin2023proteingym}. These works motivate our use of masked-reconstruction perplexity as a model-relative notion of nativeness. We show that perplexity corresponds to a dominant geometric axis in embedding space that organizes entire biological groups.

\paragraph{Biological signal in pLM representations}
Recent work using tools from mechanistic interpretability, such as sparse autoencoders (SAEs), have shown that pLM representations contain latent features associated with binding sites, structural motifs, functional domains, thermostability, localization, and family-specific constraints \citep{adams2025mechanistic, simon2025interplm, silberg2025functional}. However, these studies do not examine viral proteins, leaving open whether pLM representations encode viral-specific signal.


\paragraph{Viral protein representations}
Viral proteins are among the most difficult sequences for pLMs to model \citep{Vieira2026.03.08.710389}, yet viral pLM embeddings are still widely used. pLM embeddings have been used to annotate viral proteins beyond remote homology \citep{flamholz2024large}, to improve phage-host prediction from receptor-binding proteins \citep{gonzales2023protein}, and to improve virus-host association and host-range prediction, often outperforming sequence-based baselines \citep{liu2024prediction, panigrahi2025phage, beltran2026protein, wangpnas, wang2026without}. \citet{ofer2025viruses} even showed that viral and cellular proteins are linearly separable from mean-pooled ESM2 embeddings alone. Although these works establish that viral proteins are both atypical and informative in pLM representation space, they do not explain what drives their separation from cellular proteins.


\section{Data and Methods}
\label{sec:methods}
\subsection{Experimental setting}
\label{sec:experimental_setting}

\paragraph{Protein language models}
We evaluate three ESM-family pLMs spanning over three orders of magnitude in parameter count~\citep{rives2021esm1b}. \textbf{ESM2}~\citep{lin2023esm2} is a masked language model trained on UniRef$_{50}$~\citep{suzek2015uniref} (8M, 35M, 150M, 650M, 3B, and 15B). \textbf{ESMC}~\citep{esmc2024} is the current state-of-the-art representation model in the ESM family, also trained with masked language modeling (300M, 600M, and 6B). \textbf{ESM3}~\citep{hayes2025simulating} is a multimodal generative model jointly trained over sequence, structure, and function tracks (1.4B \textsc{open}, 1.4B \textsc{small}, 7B \textsc{medium}, and 98B \textsc{large}). ESM3-\textsc{open} is the only model trained without viral sequences; all others use virus-inclusive training data.

For each model and each input sequence, we take the final-layer embedding of every amino acid residue and mean-pool these residue embeddings, excluding BOS and EOS tokens. This gives one sequence embedding per protein.

\paragraph{Masked-reconstruction perplexity}
We use masked-reconstruction perplexity (PPL) as a sequence-level score. For each sequence $\mathbf{x}$, we mask a fraction $p{=}0.15$ of residue positions (excluding BOS/EOS), then compute the log-likelihood of masked tokens. The per-sequence perplexity is
\begin{equation}
  \mathrm{PPL}(\mathbf{x})
  \;=\;
  \exp\!\left(
    \frac{1}{|\mathcal{M}|}\!\sum_{i\in\mathcal{M}}
      -\log p_{\theta}\!\left(x_i \mid \mathbf{x}_{\setminus\mathcal{M}}\right)
  \right),
\end{equation}
\label{eq:ppl}
where $\mathcal{M}$ is the set of masked positions and $p_{\theta}$ denotes the conditional probability distribution over amino acids at each position. Results are averaged over three independent mask samples per sequence. For ESM3, PPL is computed with the model provided sequence input only.
This directly matches the masked-reconstruction objective used to train ESM2 and ESMC, and is aligned with the sequence-track component of ESM3's masked denoising objective.
\subsection{Sequence datasets}
\label{sec:data}

\paragraph{Pretraining dataset coverage (\cref{tab:coverage_asymmetry})}
To measure the data imbalance between cellular and viral proteins in ESM pretraining data, we count UniRef$_{50}$ cluster representatives by biological group, the same clustering level used to pretrain ESM2, ESMC, and ESM3~\citep{suzek2015uniref}. Query details are provided in Appendix \cref{app:data_queries}.

\paragraph{Multi-group biological dataset (\cref{fig:nativeness_axis,fig:family_nativization}).}
To evaluate these models, we gather ten biological groups spanning the tree of life: six cellular groups (bacteria, archaea, plants, fungi, insects, and human non-viral proteins) and four viral groups (bacterial (bacteriophage), human, plant, and invertebrate viruses). Cellular groups are drawn manually from Swiss-Prot~\citep{uniprot2025}. Non-human viral groups are assembled using host and lineage annotations: plant and invertebrate viruses are selected with UniProt virus-host taxonomy filters, while bacteriophages are identified using phage-relevant viral lineages and organism-name matching. The human-virus group (\cref{sec:human_dataset}) is a curated reference set defined for the human classification dataset. Full query strings, shared length/composition/deduplication filters, and per-group post-filter sizes are in Appendix \cref{app:data_pools}.

\paragraph{Biologically meaningless controls (\cref{fig:nativeness_axis})}
We generate three biologically meaningless sequence controls: \emph{position-shuffled cellular sequences}, \emph{position-shuffled viral sequences}, and \emph{i.i.d.\ uniform random sequences}. \emph{Position-shuffled cellular sequences} and \emph{position-shuffled viral sequences} are random permutations of the residue positions of the corresponding biological pool, preserving length and per-sequence amino acid composition while destroying positional structure. \emph{i.i.d.\ uniform random sequences} are drawn position-wise from the uniform distribution over the $20$ standard amino acids and length-matched to the viral pool.

\paragraph{Human viral/cellular classification dataset (\cref{fig:scaling_divergence})}
\label{sec:human_dataset}
The human viral group is a curated set of $5{,}203$ human-infecting viral proteins spanning $32$ annotated viral families, assembled from Swiss-Prot~\citep{uniprot2025} and \texttt{NP\_} entries from NCBI Virus~\citep{brister2015ncbivirus,goldfarb2025refseq}. The cellular group is a length-decile-stratified random multi-kingdom sample of $5{,}197$ sequences drawn from the cellular fraction of Swiss-Prot, matched in size to the human viral group. To prevent evaluation leakage from sequence homology, we pool all $10{,}400$ sequences and cluster them with MMseqs2~\citep{steinegger2017mmseqs2} at the standard $30\%$ identity and $80\%$ bidirectional coverage. Whole clusters are assigned to train, validation, and test at a $60/20/20$ split. All linear probe and zero-shot scores (defined in \cref{sec:analysis_methods}) reported in \cref{fig:scaling_divergence} are computed on the held-out test split. Full query strings, shared length/composition filters, and family composition are in Appendix \cref{app:viral_diversity}.

\subsection{Analysis methods}
\label{sec:analysis_methods}

\paragraph{Embedding geometry}



For each model, we pool the sequence embeddings of all biological groups (cellular and viral) and biologically meaningless controls into a single matrix and apply principal component analysis (PCA) jointly. We examine PC$_1$, the direction that explains the largest variance in the embedding space, to assess whether viral separation concentrates along a single dominant axis.

\paragraph{Per-family scaling analysis (\cref{fig:family_nativization})}
To measure how scale changes viral nativeness, we compute, for each human viral family, the fraction of sequences with $\mathrm{PPL}{<}5$. We use this fixed threshold across all ESMC model sizes because cellular groups already fall below it at smaller model scales. We restrict the analysis to families with at least $50$ sequences and plot the eight families with the highest native-like fraction at ESMC-$6$B.

\paragraph{Embedding linear probe}
For each pLM, we standardize the train-split mean-pooled embeddings and fit an $\ell_2$-regularized logistic regression head to predict the viral/cellular label. Linear probes are trained on the train split and AUC-ROC is reported on the held-out test split.

\paragraph{Zero-shot perplexity classifier}
\label{sec:zero}
We use the negated per-sequence perplexity, $s(\mathbf{x}){=}{-}\mathrm{PPL}(\mathbf{x})$, as a viral-group score and report its AUC-ROC on the same test split. To make AUC values directly comparable to the embedding linear probe regardless of which group has higher mean perplexity, we report $\max(\mathrm{AUC},\,1{-}\mathrm{AUC})$.

\paragraph{Baselines}
\label{sec:baselines}
We use three logistic-regression baselines on the same split: (i) sequence length (1 feature), (ii) amino acid composition (20 features), and (iii) dipeptide composition over adjacent residue pairs (400 features). The maximum AUC across these defines the \emph{baseline ceiling} shown in \cref{fig:scaling_divergence}.

\section{Results}


\begin{figure*}[t]
  \centering
  \includegraphics[width=0.95\textwidth]{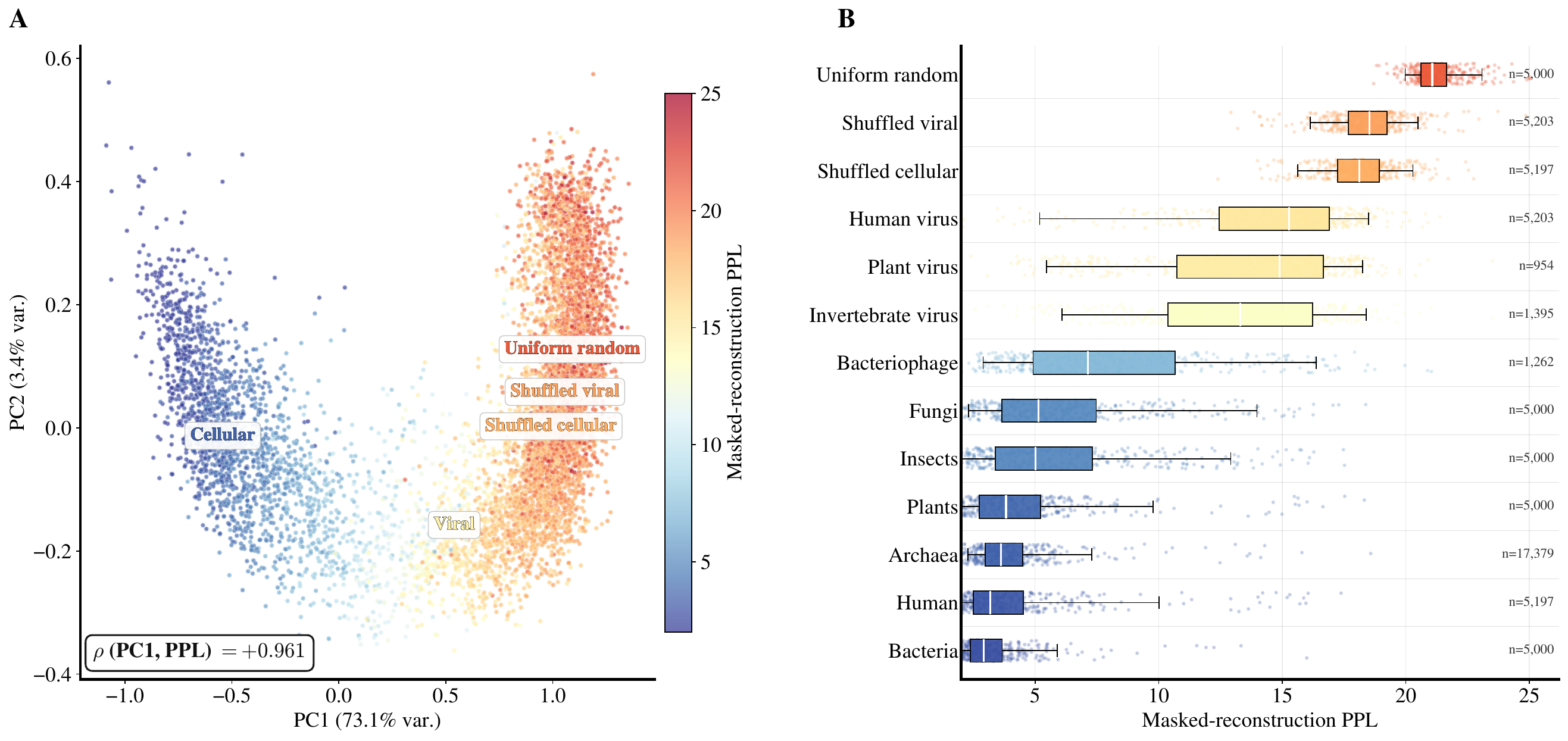}
    \caption{\textbf{A dominant nativeness axis organizes pLM representation
    space across the tree of life.}
    \textbf{(A)} PCA of ESMC-600M mean-pooled embeddings over ten biological groups and three biologically meaningless controls (\cref{sec:data}). Points correspond to individual sequences from the ten biological groups, while the overlaid labels show higher-level category centroids: cellular, viral, shuffled, and random. Each label is placed at the median coordinate of its category in (PC$_1$, PC$_2$) space and colored by that category's mean PPL. Individual points are colored by masked-reconstruction perplexity. Sequences order from well-modeled cellular proteins (low PPL, left), through the viral region, to out-of-distribution shuffled and random sequences (high PPL, right). PC$_1$ explains $73.1\%$ of variance and is strongly aligned with PPL ($\rho{=}{+}0.961$).
    \textbf{(B)} The same ordering is visible in the group-wise perplexity distributions: cellular groups have the lowest PPL, viral groups are shifted upward, and shuffled/random controls occupy the highest-PPL range. Per-group ESMC-600M masked-reconstruction perplexity ($15\%$ mask) is shown as overlaid box and strip plots, ordered by median, with $n$ reported for each group.}
  \label{fig:nativeness_axis}
\end{figure*}

\begin{table}[!b]
  \caption{\textbf{Pretraining data is strongly dominated by cellular proteins.}
    UniRef$_{50}$ cluster counts per biological group. ESM2, ESMC, and ESM3 are all pretrained at this clustering level.}
  \label{tab:coverage_asymmetry}
  \vskip 0.05in
  \begin{center}
  \begin{small}
  \begin{tabular}{lr}
  \toprule
  Biological group          & UniRef$_{50}$ clusters \\
  \midrule
  \multicolumn{2}{l}{\emph{Cellular groups}} \\
  Bacteria & $30.7\,\mathrm{M}$ \\
  Archaea & $864.9\,\mathrm{k}$ \\
  Plants & $6.0\,\mathrm{M}$ \\
  Fungi & $6.3\,\mathrm{M}$ \\
  Insects & $2.4\,\mathrm{M}$ \\
  \addlinespace
  \multicolumn{2}{l}{\emph{Viral groups}} \\
  Bacteriophage & $266.1\,\mathrm{k}$ \\
  Plant viruses & $3.4\,\mathrm{k}$ \\
  Invertebrate viruses & $6.6\,\mathrm{k}$ \\
  Human viruses (18 fam.) & $114.3\,\mathrm{k}$ \\
  \midrule
  \textbf{All cellular} & $46.3\,\mathrm{M}$ \\
  \textbf{All viral} & $390.3\,\mathrm{k}$ \\
  \midrule
  \textbf{Ratio cellular\,/\,viral} & $\mathbf{119\times}$ \\
  \bottomrule
  \end{tabular}
  \end{small}
  \end{center}
  \vskip -0.1in
\end{table}

\subsection{A dominant nativeness axis organizes pLM representation space}
\label{sec:nativeness_axis}
Viral proteins occupy a displaced yet structured region of pLM representation space. In ESMC-$600$M, PC$_1$ explains $73.1\%$ of the variance of the pooled embedding distribution (\cref{fig:nativeness_axis}A) and closely tracks masked-reconstruction perplexity (Spearman $\rho{=}{+}0.961$). This axis orders sequences from well-reconstructed cellular proteins, through viral proteins, to poorly reconstructed biologically meaningless controls, including shuffled and random sequences. The alignment also holds within individual groups: PCA refits on each group alone recover the same PPL-aligned PC$_1$ (Appendix \cref{tab:within_group_pca}). PC$_1$ thus captures a continuum from in-distribution to out-of-distribution sequences, with viral proteins occupying an intermediate region.

We refer to PC$_1$ as the \emph{nativeness axis}. A sequence is native to a pLM to the extent that it matches the statistical patterns learned during pretraining.

The same sequence ordering is visible when sequences are grouped by biological group (\cref{fig:nativeness_axis}B). Cellular groups cluster at low masked perplexity, whereas viral groups are systematically shifted upward. Bacteriophage proteins are closest to cellular proteins, whereas invertebrate, plant, and human viruses are progressively more displaced. Biologically meaningless controls occupy the high-perplexity end of the same ordering. Viral proteins are therefore not extreme outliers in pLM space; they are less native than cellular proteins under the learned prior, yet more structured than biologically meaningless sequences.



The nativeness axis generalizes across ESM families (Appendix \cref{fig:appfig_pca_ppl_esm2_esm3}): PC$_1$ explains $54.3\%$ (ESM2-650M) and $67.3\%$ (ESM3-\textsc{open}) of the variance and remains strongly correlated with PPL ($\rho{=}{+}0.926$, $+0.935$). This pattern is therefore not specific to one model architecture or training setup. It also extends beyond the masked-LM objective: an autoregressive (ProGen2) and a
discrete-diffusion (EvoDiff) pLM show the same PC$_1\!\approx\!$PPL alignment
($\rho=+0.90$ and $+0.95$) and five-tier ordering (Appendix~\ref{app:crossarch}).


This axis likely reflects both data imbalance and evolutionary constraints. ESM pretraining is dominated by cellular proteins ($\sim 119{:}1$; \cref{tab:coverage_asymmetry}), biasing the model toward them, while viral proteins are shaped by distinct evolutionary pressures. Viral sequences are thus less native both because they are underrepresented in training and because they follow different constraints.

A sequence-novelty control shows that the cellular-viral gap is not explained by pretraining data exposure alone. We evaluate $n{=}1{,}723$ cellular Swiss-Prot proteins released after the ESMC-$600$M checkpoint (Appendix \cref{app:postrelease_nonviral}), and therefore absent from pretraining. Their median PPL is $5.3$, which is much closer to the pre-release cellular reference ($3.2$) than to the viral reference ($15.3$). Thus, nativeness likely reflects compatibility with the cellular-dominated protein prior: newly released cellular proteins remain native-like, while viral proteins remain displaced.

Together, these results show that the pLM representation space contains a dominant reconstruction-aligned nativeness axis, PC$_1$, that accounts for much of the cellular-to-viral shift, while placing viral proteins between well-modeled cellular proteins and biologically meaningless controls.

\FloatBarrier
\subsection{Model scale contracts nativeness heterogeneously across human viral families}

\begin{figure}[!t]
  \centering
  \includegraphics[width=\columnwidth]{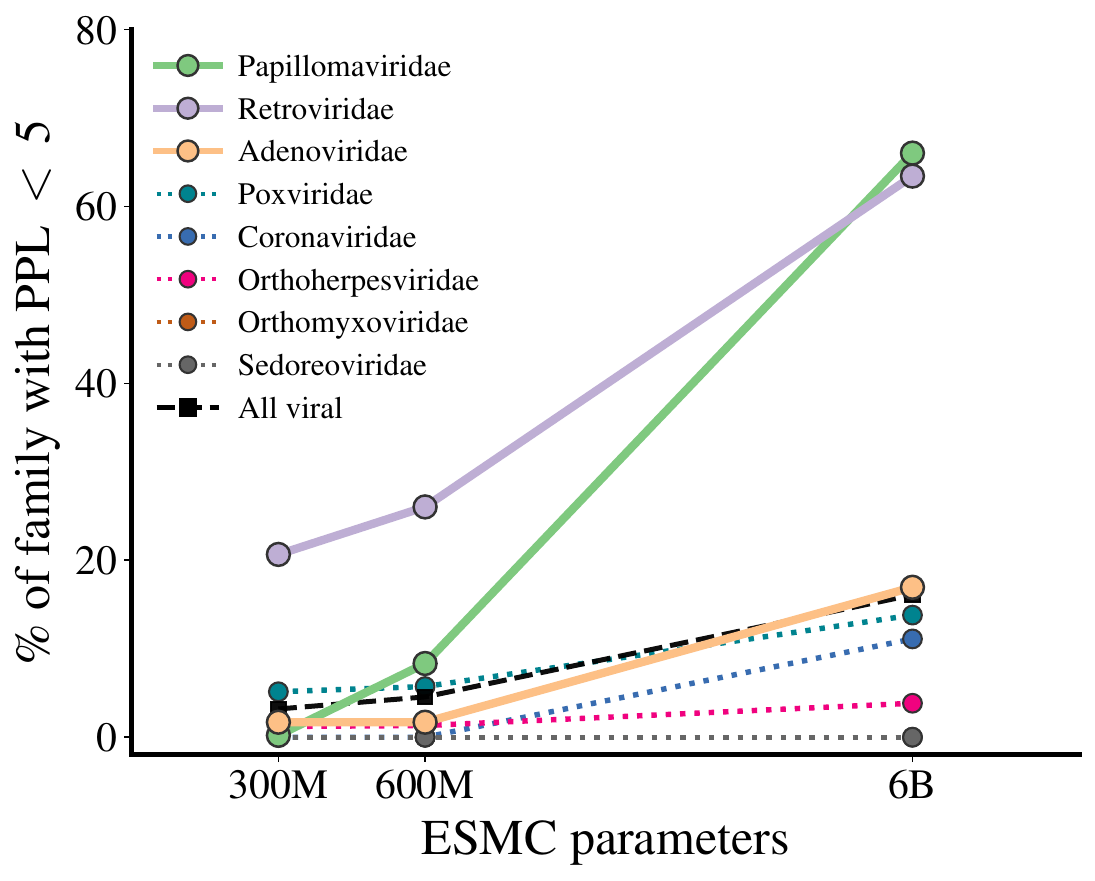}
  \caption{\textbf{Scaling contracts the nativeness axis heterogeneously
    across human viral families.}
    The fraction of sequences within each viral family whose masked-reconstruction perplexity falls
    below the native-like threshold $\mathrm{PPL}{<}5$, as a function of ESMC parameter count.
    The top three families by ESMC-6B nativization rate are drawn with a solid line whereas the rest are represented by dotted lines. The mean across all human viral sequences is represented by the dashed grey line.}
  \label{fig:family_nativization}
\end{figure}
Model scale improves viral nativeness, but not uniformly across viral biological groups. In Section 4.1, we showed that viral proteins as a whole occupy a shifted region along the nativeness axis. We now refine this analysis within a single biological group, focusing on human viruses, to examine how scaling acts across finer-grained viral families.

On average, scaling makes human viral proteins only slightly more native-like, but the effect differs strongly across families  (\cref{fig:family_nativization}). Aggregating across all human viral sequences, the fraction of native-like viral proteins increases from $\sim5\%$ at 300M to $\sim17\%$ at 6B. However, Papillomaviridae and Retroviridae gain roughly $60\%$ with scale, whereas Orthomyxoviridae, Orthoherpesviridae, and Sedoreoviridae remain mostly outside the native-like region even at 6B. This pattern suggests that scaling reduces reconstruction difficulty primarily for families already closer to the learned protein prior, while leaving more displaced families poorly reconstructed. This is consistent with the broader scaling picture, where larger models achieve lower reconstruction loss and learn richer representations \citep{kaplan2020scaling, hoffmann2022training, rives2021biological, lin2023evolutionary}, but here the contraction of the viral shift is selective rather than uniform. More specifically, recent work on rare-task retention argues that larger models are better able to learn infrequent or complex components of the training distribution because reduced representational interference allows rare-task features to accumulate rather than being overwritten by frequent tasks \citep{huang2026largermodelslearnmore}. In our setting, this provides a possible explanation for why scale can move some underrepresented viral families toward the native-like region, while leaving more distributionally distant families persistently displaced.

What may distinguish the families that become native-like is whether their proteins have cellular homologs, and Retroviridae are a clear example. Many retroviral proteins have cellular counterparts: reverse transcriptases appear in both retroviruses and eukaryotic retrotransposons \citep{menendezarias2017viral,lescot2016reverse}, and LTR retrotransposons share key structural features with retroviruses \citep{eickbush2008diversity}. This overlap means retroviral protein domains are already present in the cellular training distribution, making Retroviridae more compatible with a cellular-trained pLM.


\FloatBarrier

\begin{figure*}[t]
  \centering
  \includegraphics[width=0.98\textwidth]{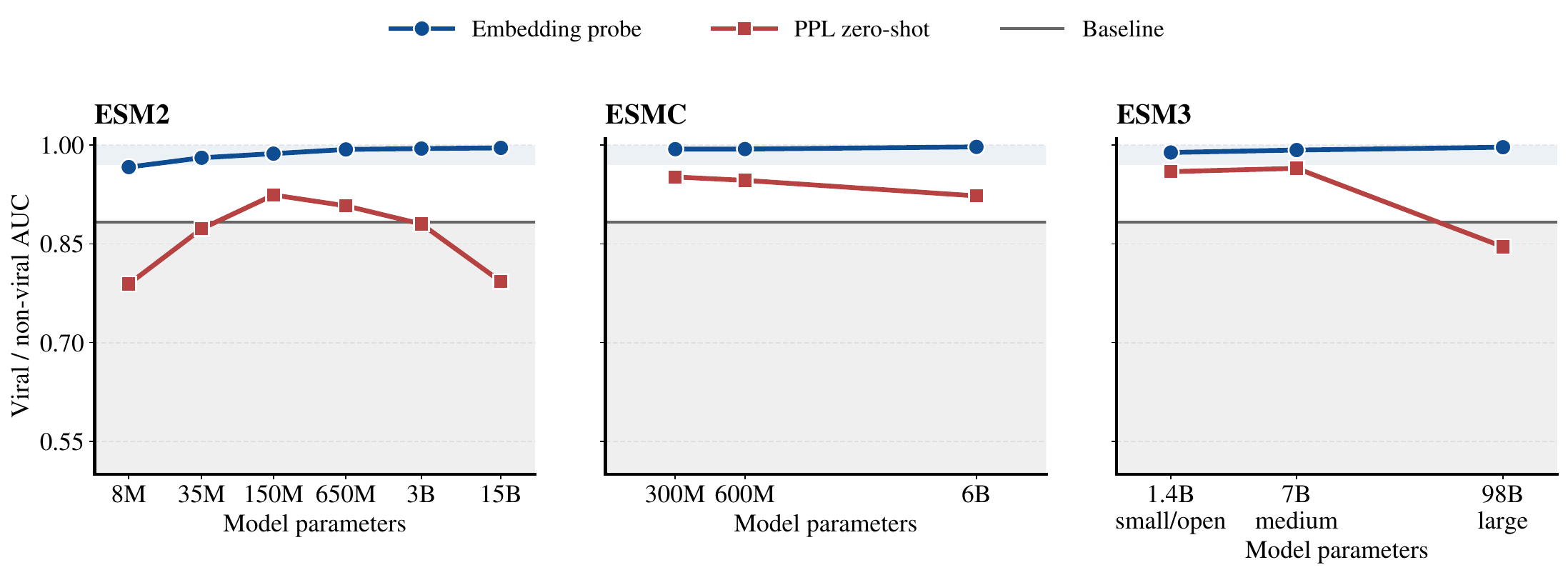}
  \caption{\textbf{Embedding linear probes capture more than perplexity: probe AUC
    scales to ceiling while PPL-based zero-shot classification does not.}
    For each ESM pLM family, we report human viral vs.\ cellular AUC-ROC on the human
    viral/cellular classification dataset (\cref{sec:data}) using two readouts of the same model: a linear probe (logistic regression) on
    mean-pooled embeddings (blue circles) and a PPL-based zero-shot
    classifier (red squares, $\mathrm{score}{=}{-}\mathrm{PPL}$). The blue band at the top
    marks the near-ceiling region $\mathrm{AUC}{\in}[0.97,1.00]$. The grey
    \emph{shallow-feature region} at the bottom spans AUC values attainable
    from raw sequence statistics alone. Its upper edge is the
    \emph{best baseline} (\cref{sec:baselines}).}
  \label{fig:scaling_divergence}
\end{figure*}

\subsection{Viral identity is linearly encoded beyond reconstruction difficulty}

Next we ask whether pLM embeddings capture biologically meaningful viral signal, or whether viral separability is simply driven by reconstruction difficulty.
For each model, we evaluate three types of classifiers on the same held-out human viral/cellular test split (\cref{sec:human_dataset}): a linear probe trained on mean-pooled embeddings, a zero-shot classifier using masked-reconstruction perplexity (PPL) alone, and the best of three baselines based on length, amino acid composition, or dipeptide composition (\cref{sec:baselines}).


\cref{fig:scaling_divergence} shows that viral identity is linearly accessible from pLM embeddings as linear probes can recover viral identity with near-ceiling accuracy across the ESM family. The linear probe AUC remains above the shallow-feature baselines at every scale and reaches the $\mathrm{AUC}\in[0.97,1.00]$ ceiling band for larger models, even under a homology-controlled train/test split.
Two additional robustness checks support that this separability reflects viral-specific signal rather than a dataset artifact. First, when the negative class is restricted to human host proteins, AUC remains at least $0.95$ for all models with more than $35$M parameters. Second, leave-one-family-out cross-validation over the $13$ viral families with at least $50$ sequences yields a mean held-out AUC above $0.96$ (Appendix~\ref{app:robustness}). This conclusion is not restricted to the ESM family: applying the same probe analysis to ProGen2 and EvoDiff yields AUCs of $0.984$ and $0.986$, respectively (Appendix~\ref{app:crossarch}).

This separability is not explained by reconstruction difficulty alone: the linear embedding probe and PPL-only classifier diverge with scale. PPL-based discrimination is weaker and non-monotonic: it improves at intermediate scales, but drops again for ESM2-$15$B and ESM3-\textsc{large}. This drop is consistent with scale making some viral proteins easier to reconstruct, moving them into the low-PPL, more native-like region and reducing the viral/cellular separation available from PPL alone. In contrast, the embedding linear probe remains near ceiling at these same scales. Thus, larger models can make some viral proteins look more native under the reconstruction objective while still preserving a linearly accessible viral signal in the embeddings.

The gap is sharpest in the low-false-positive setting relevant for sequence screening, where the embedding probe far exceeds the PPL-only classifiers across all model families and scales (Appendix~\cref{fig:appfig_tpr_low_fpr}). At $1\%$ false positive rate (FPR), the embedding linear probe reaches $88.3\%$ true positive rate (TPR) for ESM2-$15$B, $96.7\%$ for ESMC-$6$B, and $90.6\%$ for ESM3-large, compared with only $26.9\%$, $39.2\%$, and $36.1\%$ for the PPL-only classifier.
At $0.1\%$ FPR, the linear probe improves from $6.2\%$ TPR for ESM2-$8$M to $55.4\%$ for ESM2-$15$B, and from $47.9\%$ for ESMC-$300$M to $83.4\%$ for ESMC-$6$B.
Therefore, scaling improves the practical sequence screening value of the embedding representation, even when PPL alone becomes less reliable as a separator.

Finally, the linear probe also exceeds the best baseline (dipeptide composition over adjacent residue pairs), indicating that it is not merely exploiting simple sequence-level statistics.
Together, these results support our third contribution: pLM embeddings carry a residual viral signal beyond masked-reconstruction perplexity and shallow sequence features.

\section{Discussion}

\paragraph{Viral proteins remain displaced relative to a cellular-dominated prior}
ESM3-\textsc{open} was trained without viral sequences, yet it still exhibits a dominant nativeness axis aligned with masked-reconstruction perplexity, similar to the other ESM models, and viral proteins remain linearly separable from cellular proteins in embedding space. This suggests that the displaced position of viral proteins in representation space does not require direct viral exposure during pretraining. More broadly, it supports that viral proteins are positioned relative to a cellular-dominated protein prior: whether viral data are absent in pretraining, as in ESM3-\textsc{open}, or present but strongly underrepresented, as in other ESM models, viral proteins remain in a shifted region rather than being fully absorbed into the native protein manifold.

This framing gives nativeness a practical interpretation. Low nativeness may indicate that pLMs capture the constraints governing a protein family less well. Consistent with this, recent viral mutation-effect benchmarks \citep{Gurev2025.08.04.668549} suggest that sequence-only pLMs are less reliable on viral proteins than on standard, mostly non-viral benchmarks. Nativeness may therefore be useful as a \emph{diagnostic tool}: low-nativeness families may require more cautious zero-shot predictions.

\paragraph{Implications for biosecurity}

Our results suggest that pLM embeddings may carry information that is relevant for viral screening and could complement traditional homology-based methods. Homology screening primarily detects sequence similarity to known proteins, whereas embedding-based approaches may capture broader functional and biophysical signals beyond direct similarity \citep{abel2026beyond,wittmann2025strengthening}. While we do not evaluate a deployed screening system here, these findings raise the possibility that pLM-based methods may eventually help detect engineered or highly diverged variants that are less easily captured by standard homology-based filters \citep{wittmann2025strengthening,wang2026without}.

\paragraph{Limitations} Our main analysis focuses on the ESM family of protein
language models. In Appendix~\ref{app:crossarch} we provide preliminary evidence that
the nativeness axis ($\rho(\mathrm{PC}_1,\mathrm{PPL})=+0.90$ and $+0.95$) and the
residual linear viral signal (probe AUC $0.984$ and $0.986$) also emerge in a
decoder-only autoregressive model (ProGen2) and a discrete-diffusion model (EvoDiff);
the heterogeneous per-family scaling reproduces as well, though with an
objective-dependent family ranking. A systematic survey across architectures and scales
remains future work.

\paragraph{Future work}
A natural next step is to fine-tune these models on viral sequences and study the consequences for the nativeness axis. Early results show that fine-tuning lowers the perplexity on viral sequences and moves them towards the cellular/native region without degrading linear probe classification performance.

A second direction is to understand the mathematical and statistical origin of the nativeness axis. Our results suggest that masked pLMs may develop a dominant direction that tracks model fit, but the mechanism that produces this geometry is unclear. Formalizing it would help clarify whether the PC$_1$--PPL alignment is an incidental empirical feature of ESM models, a consequence of high-dimensional mixture geometry, or a more general property of masked sequence models trained on imbalanced biological data.

A third direction is cross-domain comparison, such as multilingual language models. Underrepresented languages and dialects may provide an analog of viral proteins in pLMs: both are valid structured sequences, but they are under-sampled and may be governed by partly different statistical regularities from the dominant training distribution. Studying whether multilingual LMs exhibit a comparable low-density or high-loss axis could test whether the nativeness-axis phenomenon is specific to proteins or a broader property of large masked sequence models.

\label{author info}


\section*{Impact Statement}

This work has potential positive and negative societal implications. On the positive side, understanding how pLMs represent viral proteins may improve evaluation and screening methods for biological foundation models. On the negative side, methods that better distinguish viral-like from cellular-like sequences could be misused in dual-use settings. We therefore view this work as contributing to safer biological model evaluation rather than capability deployment.

\section*{Code Availability}

Code for reproducing the main experiments, including embedding extraction,
masked-reconstruction perplexity computation, and analysis scripts, is available at
\href{https://github.com/MisteFr/viral-proteins-plms}{\texttt{github.com/MisteFr/viral-proteins-plms}}.

\nocite{langley00}

\bibliography{example_paper}
\bibliographystyle{icml2026}

\newpage
\appendix
\onecolumn

 \section{UniProt queries for pretraining coverage}
\label{app:data_queries}

\cref{tab:coverage_asymmetry} reports UniRef$_{50}$~\citep{suzek2015uniref}
cluster counts at 50\% sequence identity, the granularity at which ESM2,
ESMC, and ESM3 are pretrained. All queries were issued on 2026-04-22
against the UniProt REST endpoint and are logged for reproducibility.

\paragraph{Cellular per-taxon counts.} Each cellular group is queried by
NCBI taxonomy ID (e.g.\ \texttt{taxonomy\_id:2} for Bacteria,
\texttt{taxonomy\_id:2157} for Archaea). Per-taxon rows are \emph{not
additive}: a single UniRef$_{50}$ cluster can contain sequences from
multiple taxa and hence be counted in more than one row.

\paragraph{Viral per-taxon counts.} The UniRef REST endpoint does not
accept host filters (\texttt{virus\_host\_id}), so host-scoped viral
groups such as ``plant-infecting viruses'' cannot be counted directly
at UniRef$_{50}$ resolution. We therefore report per-taxon viral entries
only for groups queryable by lineage: Caudoviricetes (queried as
\texttt{lineage:"Caudoviricetes"}) and an $18$-family sum of
human-infecting virus families enumerated explicitly.

\section{Multi-group biological dataset: assembly details}
\label{app:data_pools}

\paragraph{Cellular groups.} For each cellular group we issue a single
UniProt query of the form
\begin{center}
\texttt{reviewed:true AND taxonomy\_id:<ID> AND (existence:1 OR existence:2 OR existence:3) AND NOT keyword:KW-0181},
\end{center}
restricting to manually reviewed Swiss-Prot entries with experimental,
transcript, or homology evidence (PE1--PE3). The final clause excludes
proteins flagged \texttt{KW-0181} (\textsc{Complete proteome}): UniProt
attaches this keyword to every entry belonging to an organism whose
proteome has been designated a reference proteome, and these few
heavily sampled species (human, \textit{E.\ coli},
\textit{S.\ cerevisiae}, \textit{A.\ thaliana}, \ldots) each contribute
tens of thousands of reviewed Swiss-Prot entries, whereas most cellular
taxa contribute only dozens. Without the filter a per-taxon query would
therefore be dominated by the proteomes of a few model organisms rather
than sample broadly across the tree of cellular life, biasing the pooled
cellular amino acid distribution accordingly.

\paragraph{Non-human-host viruses.} These groups are queried by host
taxon:
\texttt{virus\_host\_id:33090} for plant-infecting and
\texttt{virus\_host\_id:6656 AND NOT virus\_host\_id:9606} for
invertebrate-infecting viruses.

\paragraph{Bacteriophages.} Assembled in a two-step pass: we download
all reviewed viral sequences (\texttt{taxonomy\_id:10239}) and retain
those whose lineage annotation lies within a curated list of
phage-relevant clades: Caudoviricetes, Duplodnaviria, Microviridae,
Inoviridae, Tectiviridae, Leviviridae, Cystoviridae, Corticoviridae,
Plasmaviridae, Caudovirales and organism name containing
the substring ``phage''.

\paragraph{Local filtering.} Downloaded sequences are filtered to
length $\in[50,1022]$ residues (the common pLM context window, minus
BOS/EOS), $<\!5\%$ non-standard amino acids, and
exact-sequence-deduplicated by SHA-$256$ hash. If more than $5{,}000$
sequences survive these filters we draw a uniform random subsample of
size $5{,}000$ with random seed $42$; otherwise we keep all surviving
sequences.

\paragraph{Post-filter pool sizes.} Plant viruses $954$, invertebrate
viruses $1{,}395$, bacteriophage $1{,}262$; bacteria, plants, fungi,
insects each capped at the $5{,}000$ subsample; archaea is retained in full at $17{,}379$ sequences; human viruses use
the curated $5{,}203$-sequence pool of \cref{sec:human_dataset}.

\section{Human classification dataset: assembly and composition}
\label{app:viral_diversity}

\paragraph{Viral group.} The viral positives are the union of two
sources. First, $6{,}044$ Swiss-Prot~\citep{uniprot2025} entries
returned by
\begin{center}
\texttt{reviewed:true AND virus\_host\_id:9606 AND (existence:1 OR existence:2 OR existence:3)},
\end{center}
restricting to manually reviewed entries with experimental,
transcript, or homology evidence (PE1--PE3). Second, $620$ RefSeq
\texttt{NP\_}~\citep{goldfarb2025refseq} entries manually exported
from NCBI Virus~\citep{brister2015ncbivirus} under a human-host
filter, capturing curated viral proteins not yet in Swiss-Prot.

\paragraph{Cellular group.} Cellular negatives are drawn from the
non-viral fraction of Swiss-Prot (reviewed entries from any taxon
other than viruses). Sequences are bucketed into length deciles of
the viral pool and a uniform random subsample is drawn from each
decile so that the cellular length distribution matches the viral
one.

The human-virus pool is composed of 32 viral families.
The pool also covers all seven Baltimore classes, a standard virology
taxonomy~\citep{baltimore1971expression} that groups viruses by the
form of their genome and the route used to transcribe it to mRNA:
double-stranded DNA (dsDNA), single-stranded DNA (ssDNA),
double-stranded RNA (dsRNA), positive-sense single-stranded RNA
($+$ssRNA), negative-sense single-stranded RNA ($-$ssRNA), and the two
reverse-transcribing classes ssRNA-RT (e.g.\ retroviruses) and
dsDNA-RT (e.g.\ hepadnaviruses). Therefore our dataset spans the full range of viral
replication strategies rather than a single DNA or RNA lineage.

\begin{table}[h]
  \centering
  \caption{\textbf{Family composition of the human-virus pool.}
    Per-family sequence counts and percentages after exact deduplication
    and length/composition filtering. The top five families account for
    $70.1\%$ of the pool; $230$ sequences ($4.4\%$) carry no
    family-level annotation.}
  \label{tab:human_virus_families}
  \small
  \begin{tabular}{lrr}
    \toprule
    Family & $n$ & \% \\
    \midrule
    Orthomyxoviridae     & $1{,}026$ & $19.7$ \\
    Orthoherpesviridae   & $990$     & $19.0$ \\
    Poxviridae           & $681$     & $13.1$ \\
    Retroviridae         & $484$     & $9.3$  \\
    Papillomaviridae     & $468$     & $9.0$  \\
    Sedoreoviridae       & $234$     & $4.5$  \\
    Hepadnaviridae       & $194$     & $3.7$  \\
    Adenoviridae         & $177$     & $3.4$  \\
    Paramyxoviridae      & $163$     & $3.1$  \\
    Rhabdoviridae        & $106$     & $2.0$  \\
    Filoviridae          & $80$      & $1.5$  \\
    Coronaviridae        & $54$      & $1.0$  \\
    Pneumoviridae        & $52$      & $1.0$  \\
    Kolmioviridae        & $33$      & $0.6$  \\
    Arenaviridae         & $30$      & $0.6$  \\
    Polyomaviridae       & $28$      & $0.5$  \\
    Flaviviridae         & $24$      & $0.5$  \\
    Hantaviridae         & $20$      & $0.4$  \\
    Hepeviridae          & $19$      & $0.4$  \\
    Anelloviridae        & $19$      & $0.4$  \\
    Parvoviridae         & $17$      & $0.3$  \\
    Phenuiviridae        & $15$      & $0.3$  \\
    Astroviridae         & $13$      & $0.2$  \\
    Picornaviridae       & $11$      & $0.2$  \\
    Caliciviridae        & $11$      & $0.2$  \\
    Peribunyaviridae     & $8$       & $0.2$  \\
    Nairoviridae         & $5$       & $0.1$  \\
    Togaviridae          & $4$       & $0.1$  \\
    Spinareoviridae      & $2$       & $<0.1$ \\
    Picobirnaviridae     & $2$       & $<0.1$ \\
    Circoviridae         & $2$       & $<0.1$ \\
    Tobaniviridae        & $1$       & $<0.1$ \\
    \midrule
    (unannotated)        & $230$     & $4.4$  \\
    \midrule
    \textbf{Total}       & $\mathbf{5{,}203}$ & $\mathbf{100.0}$ \\
    \bottomrule
  \end{tabular}
\end{table}

\section{Embedding extraction}
\label{app:embeddings}

For locally executed checkpoints we read the final-layer hidden states
directly from the \texttt{transformers}/\texttt{esm} forward pass. For
Forge-API-only ESM3 configurations (\textsc{small}, \textsc{medium},
\textsc{large}) we encode each sequence via
\texttt{tokens = client.encode(ESMProtein(sequence=s))} and call
\texttt{client.logits(tokens, LogitsConfig(return\_embeddings=True))},
which returns the same final-layer activations as the open weights would.
Mean-pooling over the residue axis (excluding BOS/EOS) is applied
identically to both sources.

Embedding dimensionality $d$ varies across the model registry: $320$
(ESM2-$8$M), $480$ (ESM2-$35$M), $640$ (ESM2-$150$M), $1{,}280$
(ESM2-$650$M), $2{,}560$ (ESM2-$3$B), $5{,}120$ (ESM2-$15$B); $960$
(ESMC-$300$M), $1{,}152$ (ESMC-$600$M), $2{,}560$ (ESMC-$6$B); and
$1{,}536$, $2{,}560$, $6{,}144$ for Forge ESM3 \textsc{small},
\textsc{medium}, \textsc{large}. The pooling recipe is identical across
all models.

\section{Perplexity vocabulary}
\label{app:vocab}

ESM2, ESMC, and the ESM3 sequence track all use a $33$-token sequence
vocabulary~\citep{lin2023esm2,esmc2024,hayes2025simulating} including the
$20$ standard amino acids, the $5$ IUPAC ambiguity codes
\texttt{X, B, U, Z, O}, and $8$ additional special and formatting tokens
including \texttt{\textless cls\textgreater},
\texttt{\textless pad\textgreater},
\texttt{\textless eos\textgreater}, and
\texttt{\textless mask\textgreater}.

The softmax denominator in \cref{eq:ppl} is taken over this full
vocabulary. Consequently the absolute upper bound on $\mathrm{PPL}$ is
$|\mathcal{V}|{=}33$ (uniform over all tokens). Therefore we can observe a $\mathrm{PPL}$ greater than $20$ reflecting a small leakage of probability mass onto the $13$
non-amino acid tokens. That's the case for the random uniform $\mathrm{PPL}$.

\section{Within-group PCA: PC1 is the nativeness axis }
\label{app:within_group_pca}

\cref{fig:nativeness_axis}A fits PCA on the pooled embeddings of all 13 sequence pools: ten biological groups plus three synthetic controls, corresponding to the higher-level categories cellular, viral, shuffled, and random. Because these groups have different centroids and perplexity ranges, PCA1 may be biased. TThis bias, called anisotropy bias, was documented for transformer representations
in ~\citet{ethayarajh2019contextual}. The reported
$\rho(\text{PC}_1,\text{PPL}){=}{+}0.961$ could therefore be inflated by
the cellular--viral--random group separation alone, rather than reflecting
a continuous per-sequence nativeness coordinate.

We test this directly by \emph{refitting PCA inside each group} and recomputing
$\rho(\text{PC}_k,\text{PPL})$ on that refit for $k\in\{1,2,3\}$. All
embeddings are ESMC-$600$M mean-pooled representations, as in
\cref{fig:nativeness_axis}A. \cref{tab:within_group_pca} reports the
within-group refit for every natural population and for the three
biologically meaningless controls, together with two pooled anchors (cellular-only and
viral-only) that exclude the other group entirely. We also report, for
each group, the PC on which $|\rho|$ is largest over the top-3 components.

\begin{table*}[t]
\centering
\small
\caption{Within-group PCA refit. For each group, PCA is fit on that
group's ESMC-$600$M embeddings alone, and
Spearman $\rho$ between each of the first three principal components and
masked-reconstruction PPL is reported. $\mathrm{var}_1$ is the fraction of
variance captured by PC$_1$ of the refit. The final column identifies
the PC carrying the largest $|\rho|$ with PPL.}
\label{tab:within_group_pca}
\begin{tabular}{lrrrrrc}
\toprule
Group & $n$ & $\mathrm{var}_1$ (\%) & $\rho_{\mathrm{PC}_1}$ & $\rho_{\mathrm{PC}_2}$ & $\rho_{\mathrm{PC}_3}$ & $\arg\max_k |\rho_{\mathrm{PC}_k}|$ \\
\midrule
\multicolumn{7}{l}{\emph{Biological groups}} \\
Archaea                         & 17{,}379 & 28.6 & $+0.849$ & $+0.021$ & $+0.052$ & PC$_1$ \\
Bacteria                        &  5{,}000 & 30.1 & $+0.874$ & $+0.001$ & $-0.137$ & PC$_1$ \\
Fungi                           &  5{,}000 & 51.5 & $+0.963$ & $-0.041$ & $-0.157$ & PC$_1$ \\
Insects                         &  5{,}000 & 44.5 & $+0.955$ & $-0.053$ & $-0.210$ & PC$_1$ \\
Plants                          &  5{,}000 & 38.7 & $+0.934$ & $+0.061$ & $-0.095$ & PC$_1$ \\
Human non-viral                 &  5{,}197 & 41.5 & $+0.905$ & $+0.061$ & $-0.105$ & PC$_1$ \\
Bacteriophage                   &  1{,}262 & 54.3 & $+0.945$ & $+0.138$ & $-0.037$ & PC$_1$ \\
Plant virus                     &    954   & 59.5 & $+0.893$ & $+0.047$ & $-0.061$ & PC$_1$ \\
Invertebrate virus              &  1{,}395 & 47.9 & $+0.905$ & $+0.054$ & $+0.103$ & PC$_1$ \\
Human virus                     &  5{,}203 & 50.0 & $+0.830$ & $-0.112$ & $-0.032$ & PC$_1$ \\
\midrule
\multicolumn{7}{l}{\emph{Pooled groups}} \\
Cellular-only (6 groups)        & 42{,}576 & 36.8 & $+0.910$ & $-0.144$ & $+0.055$ & PC$_1$ \\
Viral-only (4 groups)           &  8{,}814 & 56.6 & $+0.894$ & $-0.048$ & $+0.024$ & PC$_1$ \\
\midrule
\multicolumn{7}{l}{\emph{Biologically meaningless controls}} \\
Shuffled cellular               &  5{,}197 & 19.6 & $-0.111$ & $-0.398$ & $-0.097$ & PC$_2$ \\
Shuffled viral                  &  5{,}203 & 22.9 & $-0.070$ & $-0.269$ & $-0.146$ & PC$_2$ \\
Random uniform                  &  4{,}999 & 34.4 & $-0.031$ & $-0.018$ & $+0.010$ & PC$_1$ \\
\bottomrule
\end{tabular}
\end{table*}

Two observations follow.

\textbf{(i) The alignment survives every single group refit.}
Across all ten biological groups, PC$_1$ of the group refit is
the PPL-aligned axis, with $\rho(\text{PC}_1,\text{PPL})\in[{+}0.830,{+}0.963]$.
PC$_2$ and PC$_3$ carry near-zero PPL signal in every case
($|\rho|{\leq}0.21$).

\textbf{(ii) The alignment is not a generic property of PCA on
embeddings.} Fitting PCA on i.i.d.\ uniform random sequences yields a PC$_1$
that is still a real top direction of variance ($34.4\%$) but carries no
PPL signal on any of its first three components
($|\rho|{\leq}0.03$). 

\section{Post-release cellular sequence-novelty control}
\label{app:postrelease_nonviral}

The goal is to test whether the
cellular--viral perplexity gap is primarily explained by training
exposure. If exposure were the dominant factor, then cellular proteins
released after the model checkpoint, and therefore absent from that
model's pretraining data, should move toward the viral distribution. We find
that they do not: post-release cellular proteins remain much closer to
the cellular reference distribution than to viral proteins.

\paragraph{Dataset.}
We use ESMC-$600$M, released as \texttt{esmc-600m-2024-12}. Because the
checkpoint predates entries created in 2025, we query UniProtKB
Swiss-Prot~\citep{uniprot2025} for reviewed, cellular entries
(\texttt{NOT taxonomy\_id:10239}) with \texttt{date\_created} on or
after $2025$-$01$-$01$. These sequences were not available when the
released checkpoint was produced. We apply the same filters used for the
main sequence pools: length in $[50,1022]$, fewer than $5\%$
non-standard amino acids, and exact-sequence deduplication by SHA-$256$
hash (\cref{app:data_pools}). This yields $1{,}723$ post-release
cellular sequences.

\paragraph{Scoring.}
We compute masked-reconstruction perplexity with the same protocol used
throughout the paper (\cref{eq:ppl}): $15\%$ masking, three independent
mask samples per sequence, and the ESMC-$600$M sequence model. As
reference distributions, we reuse the ESMC-$600$M perplexities from the
human viral/cellular classification dataset (\cref{sec:human_dataset}),
pooling the train, validation, and test splits. This gives $5{,}197$
cellular proteins that were already present in sequence databases before
the checkpoint release, and $5{,}203$ curated human-virus proteins.

\begin{figure*}[t]
  \centering
  \includegraphics[width=0.96\textwidth]{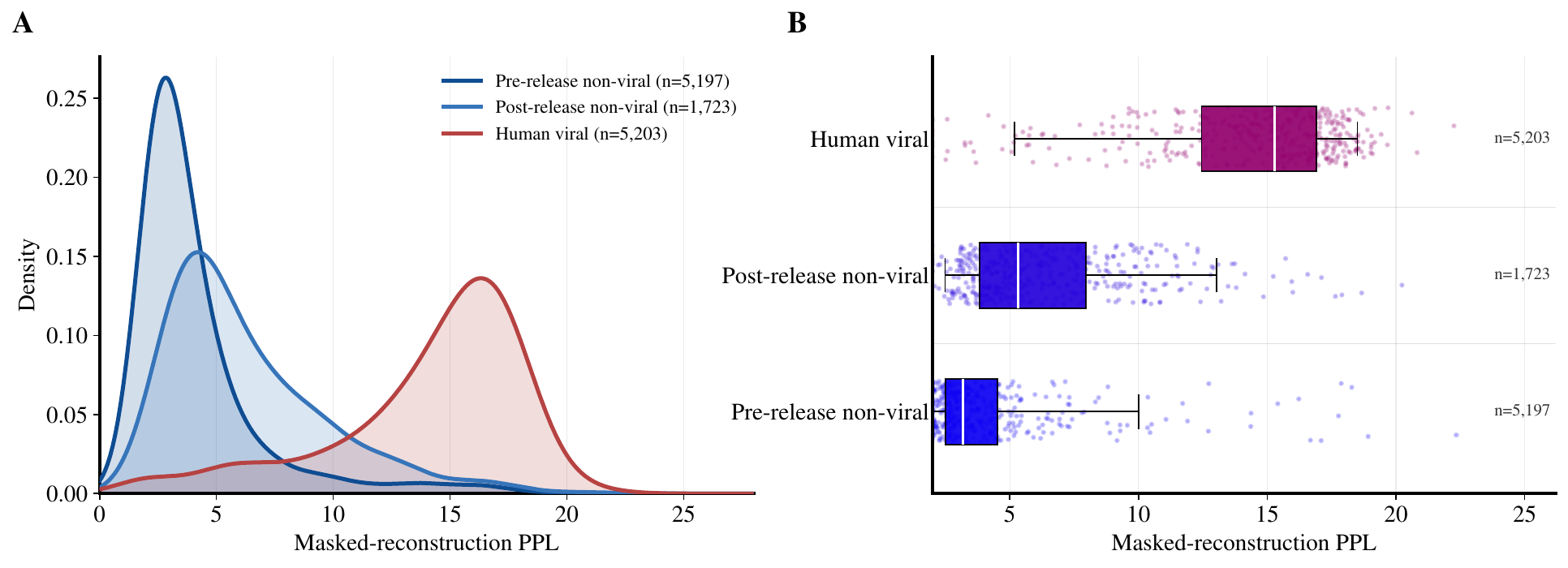}
  \caption{\textbf{Post-release cellular proteins remain native-like
    under ESMC-$600$M.}
    Masked-reconstruction perplexity for three sequence groups:
    cellular proteins already present in sequence databases before the
    checkpoint release ($n=5{,}197$), post-release cellular Swiss-Prot
    proteins created on or after $2025$-$01$-$01$ ($n=1{,}723$), and the
    curated human-virus pool from \cref{sec:human_dataset}
    ($n=5{,}203$).}
  \label{fig:appfig_postrelease_control}
\end{figure*}

\paragraph{Result.}
The post-release cellular proteins remain native-like
(\cref{fig:appfig_postrelease_control}). Median PPL is $3.2$ for the
pre-release cellular reference set, $5.3$ for the post-release
cellular set, and $15.3$ for the viral set. Thus the post-release
distribution is shifted upward relative to the pre-release cellular
reference distribution, but the shift is small compared with the viral
displacement.
cellular proteins absent from the checkpoint's pretraining data remain
far more native to the model than viral proteins, indicating that
nativeness reflects compatibility with the learned cellular-dominated
protein prior rather than raw exposure alone.

\section{ESM2 and ESM3 per-family nativization}
\label{app:family_nativization_allfamilies}

\begin{figure*}[t]
  \centering
  \includegraphics[width=0.98\textwidth]{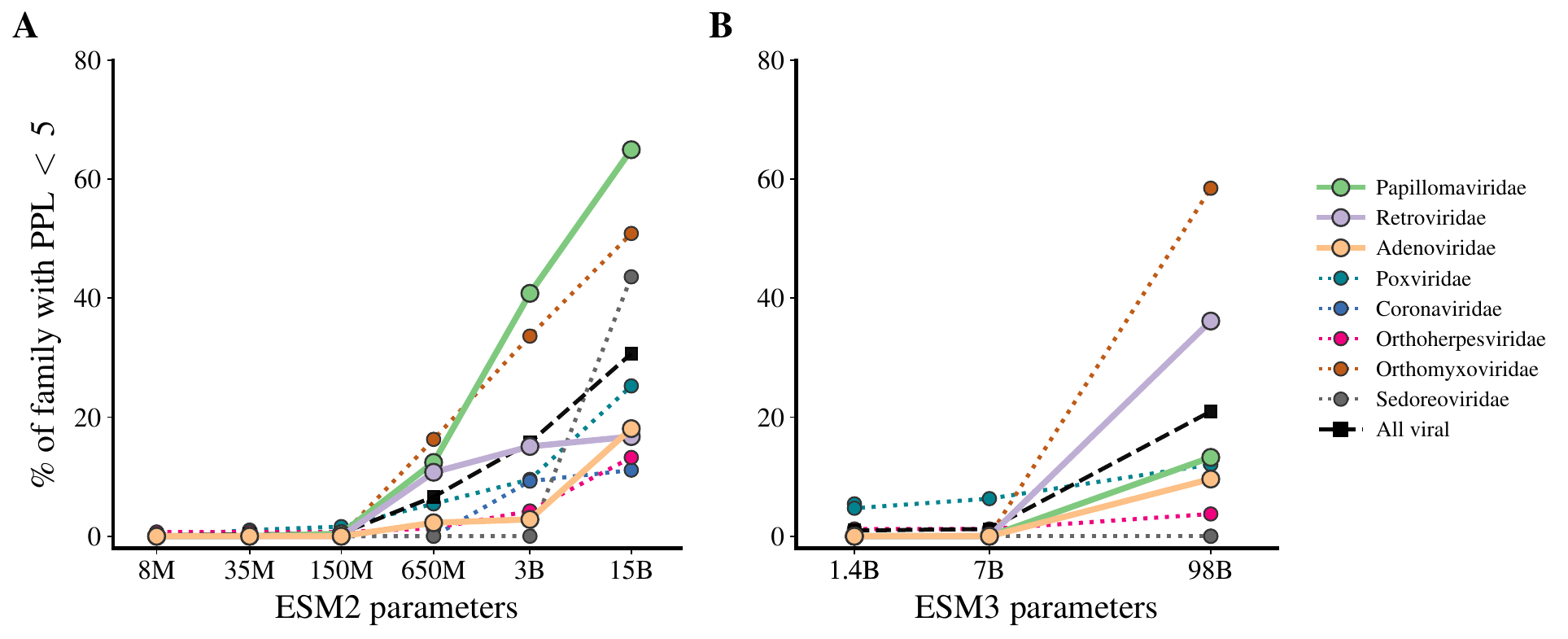}
  \caption{\textbf{Per-family nativization across ESM2 and ESM3.}
    Fraction of each viral family with masked-reconstruction PPL$<5$
    as a function of parameter count. The threshold is the same fixed
    PPL$<5$ cut used in \cref{fig:family_nativization}}
  \label{fig:appfig_family_nativization_all}
\end{figure*}

We verify that the same heterogeneous scaling is present in the other two ESM families rather than an ESMC-specific effect. \cref{fig:appfig_family_nativization_all}
We repeat the same analysis for the other two ESM architectures:
ESM2 ($8$M $\to$ $15$B, six scales) and ESM3 (\textsc{small} $\to$
\textsc{open} $\to$ \textsc{medium} $\to$ \textsc{large}, four scales). We find that the same families that become increasingly native-like under ESMC scaling (Papillomaviridae and Retroviridae) also rise under ESM2 and ESM3 scaling.

\section{ESM2 and ESM3 nativeness axis}
\label{app:pca_ppl_esm2_esm3}

\begin{figure*}[t]
  \centering
  \includegraphics[width=0.96\textwidth]{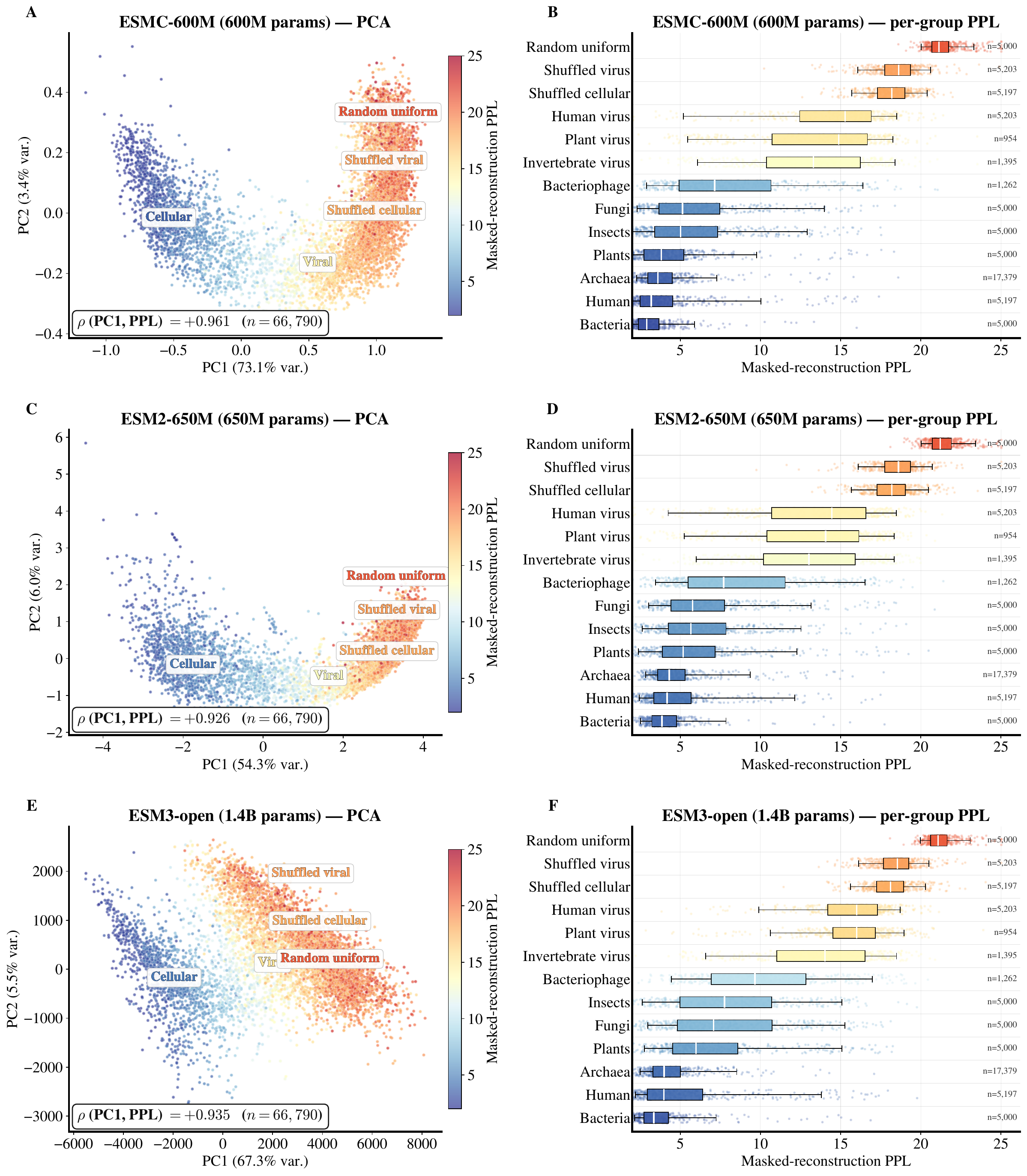}
  \caption{\textbf{The nativeness axis reproduces on ESM2-$650$M and
    ESM3-\textsc{open}.} PCA of mean-pooled embeddings, points coloured by masked-reconstruction PPL with the same
    colourmap as main \cref{fig:nativeness_axis}A.}
  \label{fig:appfig_pca_ppl_esm2_esm3}
\end{figure*}

We verify that the same \emph{nativeness axis} is present in the other two ESM families rather than an ESMC-specific effect, we repeat the PCA + PPL analysis on one checkpoint from each of ESM2 and ESM3
(\cref{fig:appfig_pca_ppl_esm2_esm3}).

We find that the qualitative claims of main
\cref{fig:nativeness_axis}A survive in both other architectures
(\cref{fig:appfig_pca_ppl_esm2_esm3}).
\textbf{(i) Low-dimensionality.} PC$_1$ alone captures $54.3\%$ of the
mean-pooled embedding variance on ESM2-$650$M and $67.3\%$ on
ESM3-\textsc{open}. \textbf{(ii) PC$_1$-PPL alignment.} The Spearman rank
correlation between PC$_1$ and masked-reconstruction perplexity is
$\rho{=}{+}0.926$ on ESM2-$650$M and $\rho{=}{+}0.935$ on
ESM3-\textsc{open}. Both values are similar to the
ESMC $\rho{=}{+}0.961$.
\textbf{(iii) Group ordering on PC$_1$.} The cellular groups sits at one
end of PC$_1$ at low PPL; the three biologically meaningless controls
(shuffled cellular, shuffled viral, i.i.d.\ uniform random) sit at the
high-PPL end; and the human-virus cloud lies between the cellular groups and the biologically meaningless tail, exactly as in main
\cref{fig:nativeness_axis}B.

\section{Low-FPR classification performance}
\label{app:low_fpr_screening}

\begin{figure}[t]
    \centering
    \includegraphics[width=\linewidth]{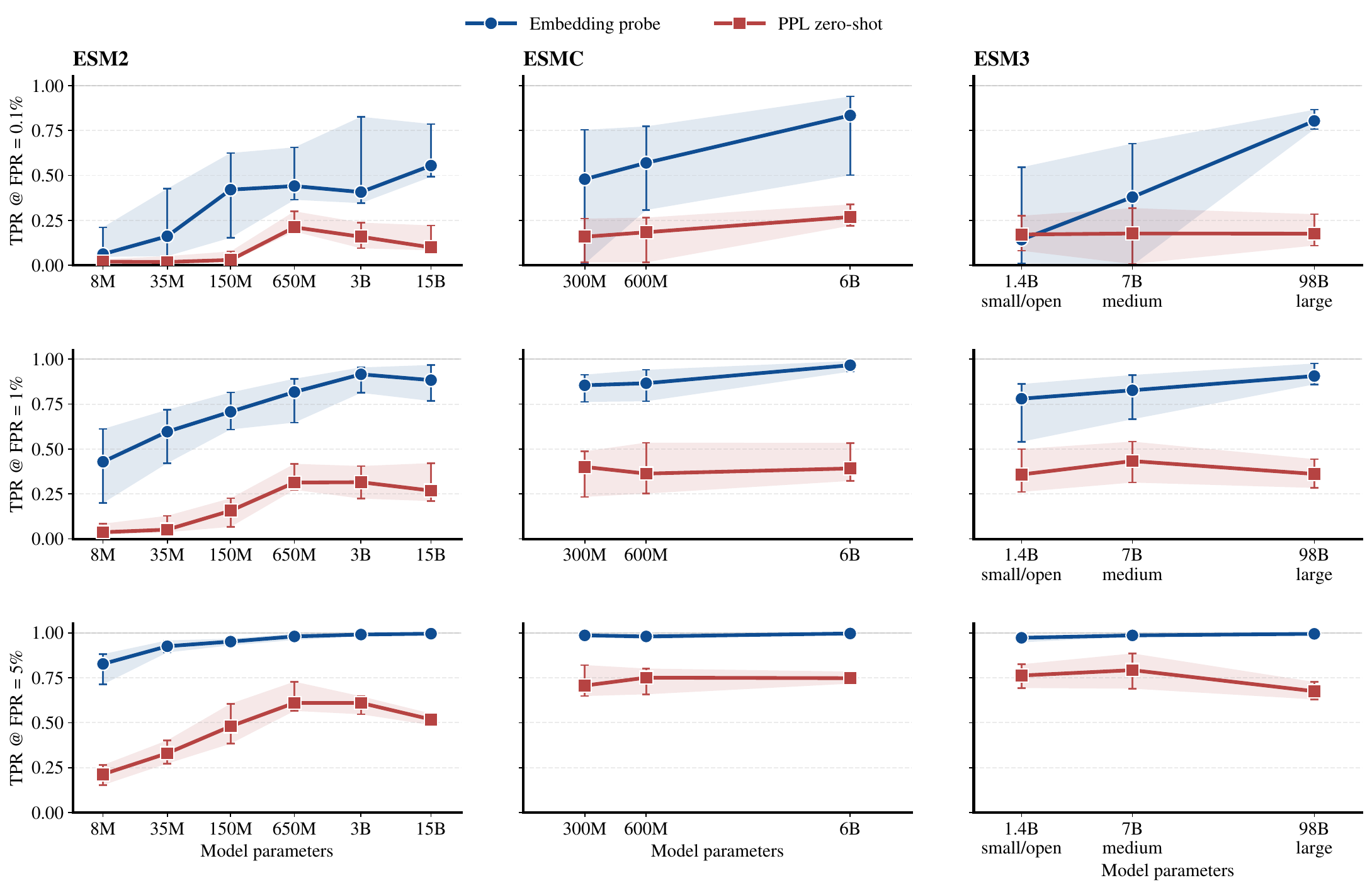}
    \caption{
    True positive rate (TPR) at low false positive rates (FPR) for embedding linear probes (blue) and PPL-based classifiers (red) across ESM2, ESMC, and ESM3 model families. Rows correspond to operating points at $0.1\%$, $1\%$, and $5\%$ FPR. Shaded regions and error bars indicate
    $95\%$ percentile bootstrap confidence intervals over $2{,}000$
    resamples of the held-out human test set ($n{=}2{,}080$). Embedding linear probes consistently achieve substantially higher TPR than PPL-based classifiers across all scales, with the gap most pronounced in the low-FPR setting relevant for screening.
    }
    \label{fig:appfig_tpr_low_fpr}
\end{figure}

We report TPR at fixed low FPR thresholds to evaluate performance in the setting relevant for screening applications. Across all model families and scales, embedding linear probes substantially outperform PPL-based classifiers. The gap is largest at very low FPR (0.1\% and 1\%), where PPL-based discrimination remains limited even for large models, while embedding linear probes approach near-ceiling performance.

These results confirm that the viral signal captured by embeddings is not only stronger in aggregate metrics such as AUC, but also more effective in practical low-false-positive settings.

\section{Probe robustness: harder negatives and held-out families}
\label{app:robustness}

We run two additional robustness checks for the linear-probe results in Section 4.3. These controls test whether the near-ceiling viral/cellular separability reflects viral-specific signal rather than a coarse dataset contrast or memorization of family-specific motifs.

\paragraph{Harder negatives.}
In the main classification dataset, cellular negatives are drawn from a length-matched multi-kingdom Swiss-Prot sample. A probe could therefore in principle exploit broad differences between human-infecting viruses and distant cellular taxa such as bacteria, plants, or fungi. To test this possibility, we re-evaluate each trained probe after restricting the held-out negative class to \emph{Homo sapiens} proteins, matching the host infected by the viral positives.

\begin{figure*}[t]
\centering
\includegraphics[width=0.78\textwidth]{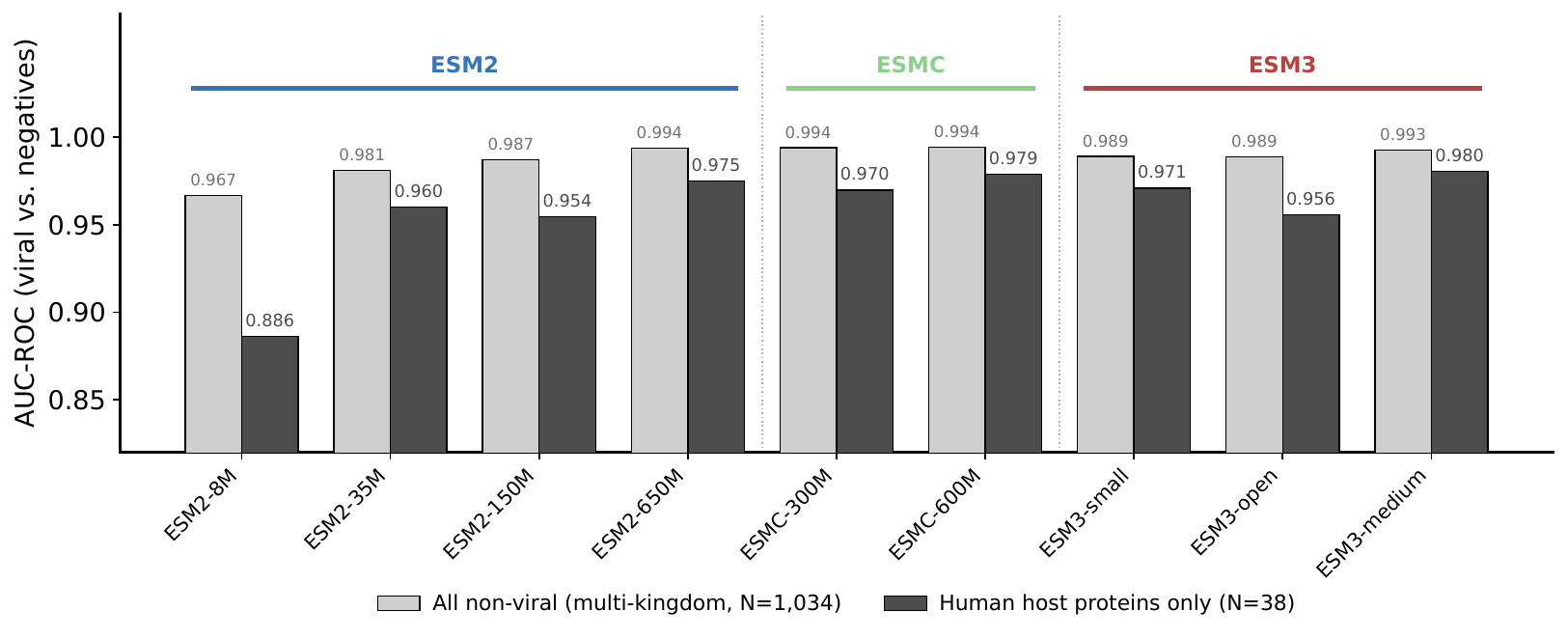}
\caption{\textbf{The probe distinguishes viral proteins from the human host itself, not only from distant cellular kingdoms.}
Per-model probe AUC-ROC on the held-out human-virus test split, evaluated against the full multi-kingdom Swiss-Prot negative pool (light, $n{=}1{,}034$) and against \emph{Homo sapiens} negatives only (dark, $n{=}38$). Human-only AUC remains at least $0.95$ for every model with at least $35$M parameters; only ESM2-8M drops, to $0.886$.}
\label{fig:human-negatives}
\end{figure*}

Probe performance is essentially unchanged (Figure~\ref{fig:human-negatives}). Human-negative AUC-ROC remains at least $0.95$ for every pLM with at least $35$M parameters, and only the smallest model, ESM2-8M, drops to $0.886$. Thus, the probe separates viral proteins from same-host cellular proteins, not only from distant cellular kingdoms. Because the held-out human-negative pool is small ($n{=}38$; the multi-kingdom test split is only ${\sim}3.7\%$ human), this estimate is noisier than the full-pool AUC.

\paragraph{Held-out families.}
We next test whether the probe relies on family-specific motifs. We run leave-one-family-out cross-validation over the $13$ viral families with at least $50$ sequences. For each split, one viral family is removed entirely from training and used only for evaluation.

\begin{figure*}[t]
\centering
\includegraphics[width=0.72\textwidth]{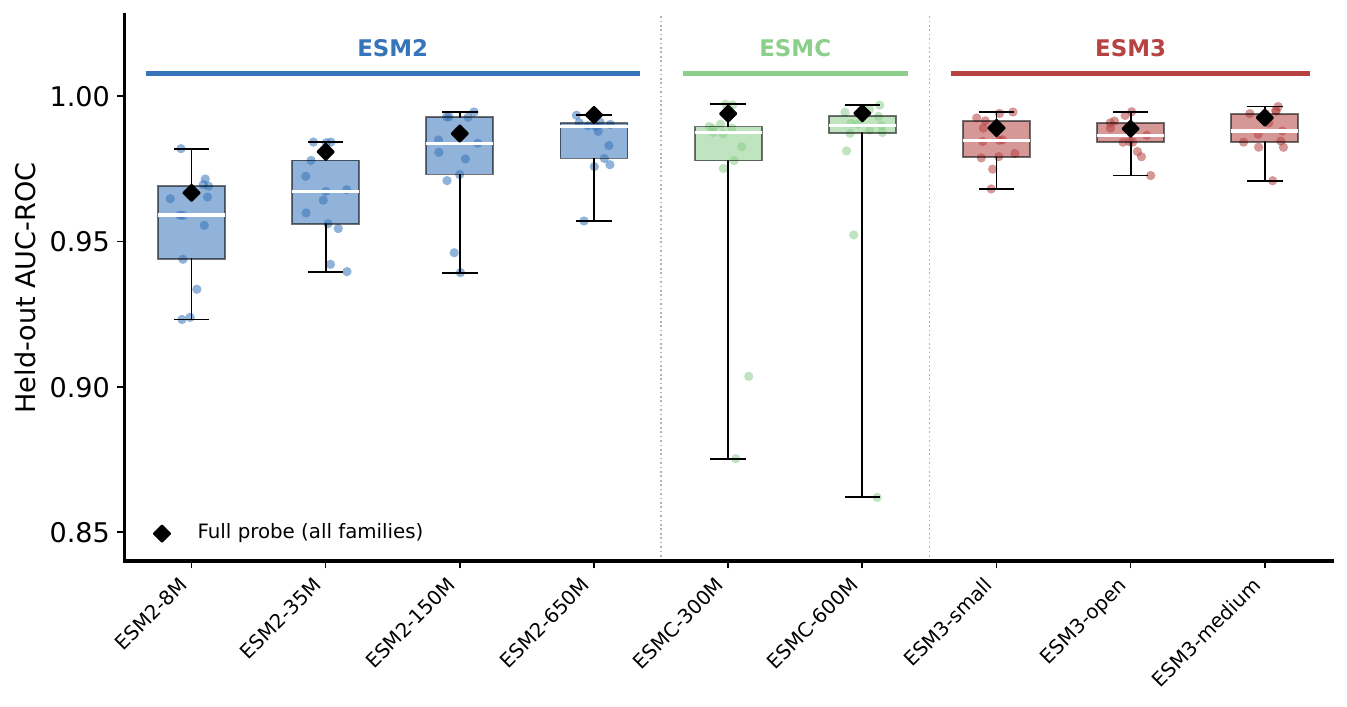}
\caption{\textbf{Held-out viral families remain separable, indicating that the probe does not simply memorize family-specific motifs.}
Leave-one-family-out cross-validation across the $13$ viral families with at least $50$ sequences. For each pLM, the box and jittered points show held-out AUC-ROC across families; the black diamond marks the model's full within-distribution probe AUC. Held-out AUC closely tracks the full probe. The lowest ESMC evaluations correspond to Retroviridae, down to $0.862$.}
\label{fig:leave-family-out}
\end{figure*}

Held-out family performance remains close to the full-probe performance (Figure~\ref{fig:leave-family-out}). The median held-out AUC-ROC is $0.984$, and only three evaluations fall below $0.92$. All three correspond to Retroviridae, consistent with the endogenous-retrovirus overlap with host genomes discussed in Section 4.2 and with Retroviridae being closest to the cellular manifold. These results indicate that the probe captures a family-general viral signal rather than memorizing family-specific sequence motifs.

\section{Cross-architecture generalization: autoregressive and diffusion pLMs}
\label{app:crossarch}

The main analysis focuses on the ESM family, whose models are trained with masked or
span-denoising reconstruction objectives. We next ask whether the same three results
hold outside this family. We evaluate two non-ESM, sequence-only generative pLMs:
\textbf{ProGen2-base} (764M parameters), a decoder-only autoregressive Transformer
\citep{nijkamp2023progen2}, and \textbf{EvoDiff OA-DM} (640M parameters), a discrete
order-agnostic diffusion model \citep{alamdari2023evodiff}. Both models are evaluated
on the same 13-group sequence pool ($n=66{,}790$) and the same held-out human
viral/cellular classification split used in the main experiments.

Perplexity has a different meaning for each objective: masked-token reconstruction for
ESM, causal next-token likelihood for ProGen2, and the order-agnostic ELBO for EvoDiff
\citep{hoogeboom2022autoregressive}. These values are therefore not comparable in
absolute scale across architectures. We only compare within-model quantities: the
alignment between PC$_1$ and perplexity, the ordering of sequence groups, and the
relative performance of embedding probes, perplexity-only classifiers, and shallow
sequence baselines.

\paragraph{Models and embeddings.}
The two added models differ from ESM in both architecture and training objective.
ProGen2-base is a decoder-only Transformer trained by left-to-right next-residue
prediction, whereas ESM models use bidirectional masked reconstruction. EvoDiff OA-DM is
a dilated-convolutional ByteNet model with an explicit diffusion-timestep input, trained
to reconstruct residues in an order-agnostic generation process rather than under a
fixed mask rate.

For each model, we extract one embedding per protein by mean-pooling residue-level
hidden states over amino-acid positions only, excluding special and padding tokens. For
ProGen2, we mean-pool the final-layer hidden states of the clean sequence after removing
BOS/EOS tokens, giving a 1536-dimensional embedding. For EvoDiff, we mean-pool the
pre-decoder ByteNet representation of the clean sequence at the fully generated timestep
$t=L$, captured before the final projection layer, giving a 1280-dimensional embedding.
PCA, linear probes, and all embedding analyses use these vectors without further
modification. Perplexity is computed using each model's native objective: exact causal
per-residue likelihood for ProGen2 and a 24-sample OA-ARDM ELBO estimate for EvoDiff.

\subsection{The nativeness axis appears to extend beyond masked language modeling}

\begin{figure*}[t]
\centering
\includegraphics[width=\textwidth]{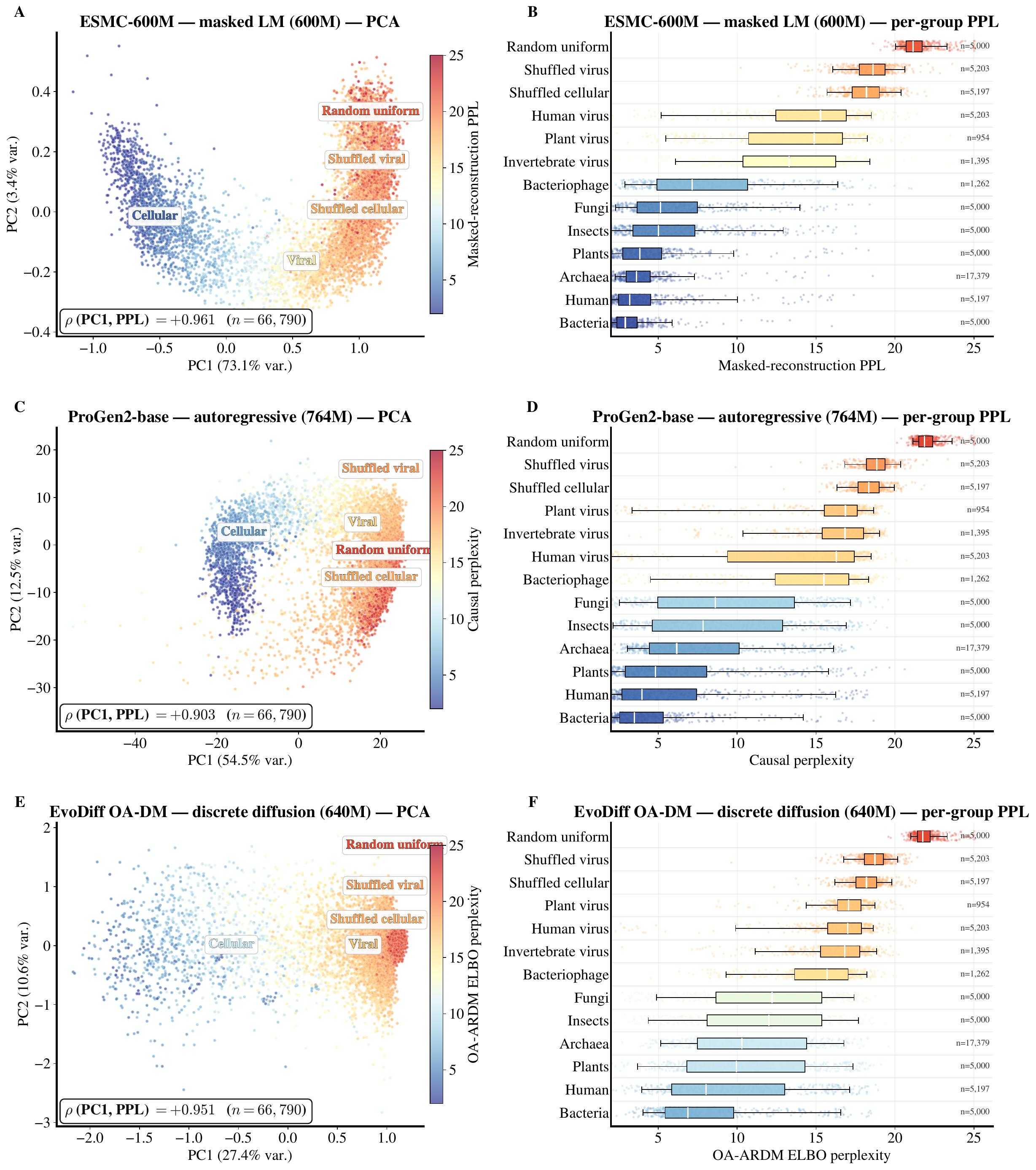}
\caption{\textbf{The nativeness axis appears beyond the masked-LM objective.}
Figure~\ref{fig:nativeness_axis} analysis repeated for three pLMs with different
training objectives, evaluated on the same 13-group sequence pool
($n=66{,}790$): ESMC-600M (masked LM; reference), ProGen2-base
(autoregressive), and EvoDiff OA-DM (order-agnostic discrete diffusion).
\emph{Left:} PCA of mean-pooled sequence embeddings, colored by each model's
native per-residue perplexity score: masked-reconstruction perplexity for ESMC,
causal perplexity for ProGen2, and the OA-ARDM ELBO-based perplexity estimate for
EvoDiff. Insets report $\rho(\mathrm{PC}_1,\mathrm{PPL})$.
\emph{Right:} group-wise perplexity distributions, shown as box and strip plots
and ordered by median. PC$_1$ is strongly aligned with perplexity in all three
models ($\rho=+0.961$, $+0.903$, and $+0.951$), and the same ordering from
cellular proteins to viral proteins to shuffled and random controls is preserved.
Because perplexity is objective-specific, absolute values should not be compared
across rows; the shared color scale is used only for visual consistency.}
\label{fig:crossarch_pca}
\end{figure*}

Figure~\ref{fig:crossarch_pca} repeats the main nativeness-axis analysis for ESMC-600M,
ProGen2-base, and EvoDiff OA-DM. The same structure appears outside the ESM family.
PC$_1$ remains strongly aligned with reconstruction difficulty, with
$\rho(\mathrm{PC}_1,\mathrm{PPL})=+0.903$ for ProGen2 and $+0.951$ for EvoDiff.
PC\textsubscript{1} explains 54.5\% of variance for ProGen2 and 27.4\% for EvoDiff, compared with 73.1\% for ESMC-600M.

The group ordering is also preserved. In both non-ESM models, cellular proteins have
the lowest perplexity, followed by human viral proteins, shuffled cellular sequences,
shuffled viral sequences, and random sequences. The group-mean perplexities follow this
ordering for ProGen2
($7.2 < 14.0 < 18.3 < 18.7 < 22.1$) and for EvoDiff
($10.5 < 16.1 < 18.1 < 18.6 < 21.9$). Thus, as in ESM, viral proteins occupy an
intermediate region between well-modeled cellular proteins and biologically meaningless
controls.

The PC$_1$--PPL alignment is strongest within the biological groups. On the shared
PC$_1$, the within-group correlations for cellular and viral proteins are
$+0.82$ and $+0.86$ for ProGen2, and $+0.91$ and $+0.71$ for EvoDiff. The weaker viral
correlation for EvoDiff suggests that the diffusion model distributes viral variation
across more directions, but the dominant axis still tracks reconstruction difficulty.
Together, these results show that the nativeness axis is not an artifact of the
masked-LM objective. It also appears in autoregressive and order-agnostic diffusion
pLMs.

\subsection{Scale contracts nativeness heterogeneously outside the ESM family, but the family ranking depends on objective}

\begin{figure*}[t]
\centering
\includegraphics[width=0.85\textwidth]{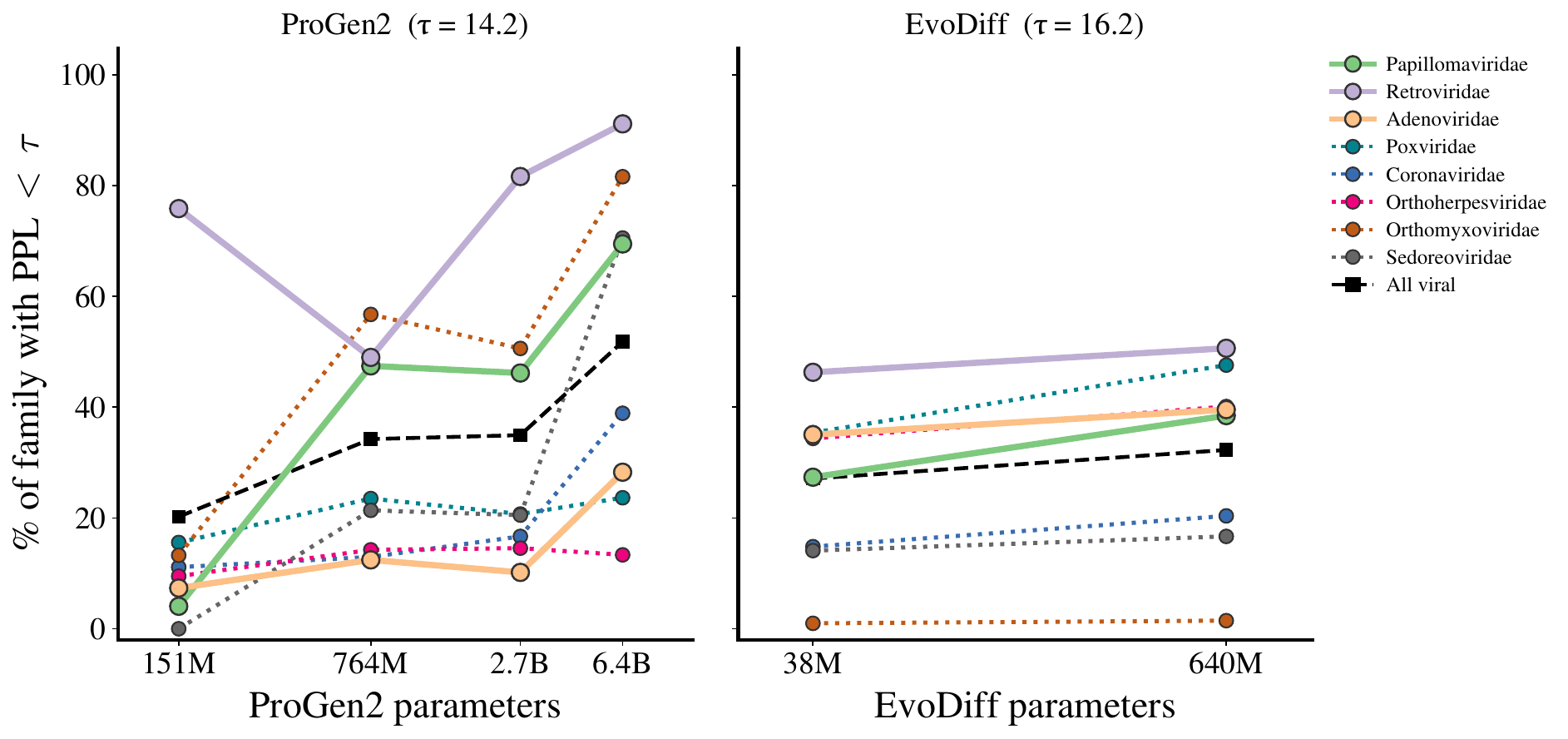}
\caption{\textbf{Scaling remains heterogeneous outside the ESM family, but the
family ranking depends on the objective.}
Fraction of each human viral family below a model-family-specific native-like
threshold $\tau$, plotted as a function of parameter count for ProGen2
(autoregressive; four scales) and EvoDiff (diffusion; two scales). For each
architecture, $\tau$ is defined as the 90th percentile of cellular perplexity at
the reference scale and is then held fixed across scales. Solid lines mark the
three families that nativize most strongly under ESMC-6B in
Figure~\ref{fig:family_nativization}; colors are matched across figures. ProGen2 shows
heterogeneous nativization, but its strongest nativizers are Orthomyxoviridae and
Sedoreoviridae, two families that remain displaced under ESMC. EvoDiff shows a
weaker viral-family response over its two available scales. Since $\tau$ is
architecture-specific, the vertical axis is not directly comparable to the
fixed $\mathrm{PPL}<5$ axis in Figure~\ref{fig:family_nativization}.}
\label{fig:crossarch_nativization}
\end{figure*}

The fixed $\mathrm{PPL}<5$ native-like threshold used in Figure~\ref{fig:family_nativization} is
not meaningful across objectives. ProGen2 already assigns cellular proteins a median
perplexity below $5$ at 764M parameters, whereas EvoDiff does not reach this scale
because its score is an ELBO-based perplexity bound. We therefore define a
model-family-specific native-like threshold $\tau$ as the 90th percentile of cellular
perplexity at the reference scale, giving $\tau=14.2$ for ProGen2-base and
$\tau=16.2$ for EvoDiff-640M. This threshold is then held fixed across scales within
each architecture. Consequently, the vertical axis of
Figure~\ref{fig:crossarch_nativization} should not be compared in absolute value with
Figure~\ref{fig:family_nativization}; only within-architecture trends and family rankings are
interpretable.

For ProGen2, evaluated across four scales from 151M to 6.4B parameters, scaling again
contracts nativeness selectively rather than uniformly. The fraction of viral sequences
below $\tau$ increases from $20\%$ at 151M to $52\%$ at 6.4B, while the cellular median
perplexity decreases from $6.5$ to $2.7$. However, the families that become native-like
are not the same as in ESMC. In ESMC, Papillomaviridae and Retroviridae move most
strongly toward the native region, whereas Orthomyxoviridae, Orthoherpesviridae, and
Sedoreoviridae remain displaced. In ProGen2, the strongest nativizers are instead
Orthomyxoviridae and Sedoreoviridae. Orthomyxoviridae drops from median PPL $17.8$ to
$1.8$, with the fraction below $\tau$ increasing from $13\%$ to $82\%$, and
Sedoreoviridae increases from $0\%$ to $71\%$ below $\tau$.

Other patterns are shared across objectives. Papillomaviridae nativizes in both ESMC
and ProGen2, increasing from $4\%$ to $69\%$ below $\tau$ in ProGen2.
Orthoherpesviridae remains displaced in both objectives, as do Adenoviridae,
Coronaviridae, and Poxviridae, whose median perplexities do not decrease with scale
($\Delta\mathrm{PPL}$ between $+0.4$ and $+1.3$). Retroviridae is the most consistent
case: it is already largely native-like at the smallest ProGen2 scale, with $76\%$ of
sequences below $\tau$, and plateaus thereafter. Thus, scaling nativizes some viral
families and not others outside the ESM family, but the detailed family ranking is
objective-dependent. The family ordering reported in Section 4.3 should
therefore be interpreted as partly ESM-specific.

EvoDiff shows a complementary pattern across its two available scales, 38M and 640M.
The cellular median perplexity decreases substantially, from $14.5$ to $8.0$, whereas
viral family medians remain nearly flat, with per-family changes between $-0.1$ and
$-0.7$. As a result, the viral--cellular gap widens rather than closes with scale.
Because only two EvoDiff scales are available, this should be interpreted as a
direction-of-effect result rather than a full scaling law. The largest viral shift is
again observed for Papillomaviridae ($\Delta=-0.7$), consistent with its behavior under
ESM.

These cross-architecture results also support the homology-based interpretation of
Retroviridae from Section 4.3. Retroviridae is the most native-like viral
family at the smallest scale of each architecture: ESMC-300M (21\% below the $\mathrm{PPL}<5$ threshold), ProGen2-151M (76\% below $\tau$), and EvoDiff-38M
(rank 1). It also remains among the most native-like families at the largest scale of
each architecture, ranking first for ProGen2-6.4B and EvoDiff-640M, and second for
ESMC-6B. Unlike families such as Papillomaviridae or Orthomyxoviridae, Retroviridae
does not require scale to become native-like; it starts near the native region. This is
consistent with the sequence-level mechanism proposed in Section 4.3:
retroviral protein domains overlap the cellular training distribution through
eukaryotic retrotransposons. The Retroviridae result is therefore not only reproduced
outside ESM, but strengthened by the fact that its nativeness is stable across training
objectives.

\subsection{Viral identity remains linearly encoded beyond reconstruction difficulty outside the ESM family}

\begin{figure*}[t]
\centering
\includegraphics[width=0.9\textwidth]{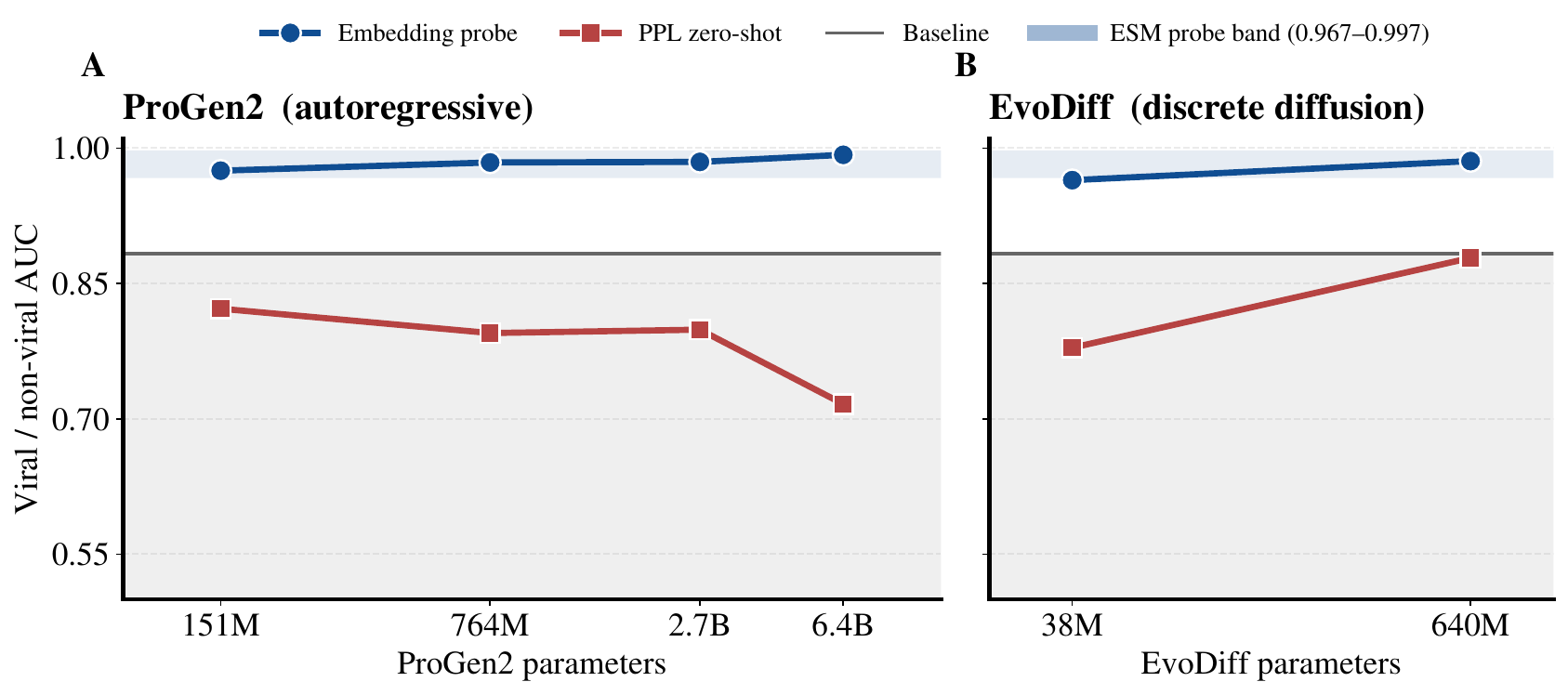}
\caption{\textbf{Embedding probes outperform perplexity-only classifiers across
scale outside the ESM family.}
Human viral/cellular AUC-ROC on the held-out test split, as a function of model
scale, for \textbf{(A)} ProGen2 (autoregressive; 151M--6.4B parameters) and
\textbf{(B)} EvoDiff OA-DM (discrete diffusion; 38M--640M parameters). Blue points
show logistic-regression probes trained on mean-pooled embeddings from each scale.
Red points show the corresponding perplexity-only zero-shot classifier
($s=-\mathrm{PPL}$, reported as $\max(\mathrm{AUC},1-\mathrm{AUC})$ as in
Section 4.3). The gray band shows the shallow-feature baseline
range, whose upper edge is the best length, amino-acid-composition, or dipeptide
baseline (AUC $=0.883$). The blue band shows the range of ESM embedding-probe AUCs
from Figure~\ref{fig:scaling_divergence} ($0.967$--$0.997$). For both non-ESM
architectures, the embedding probe is near ceiling and lies within the ESM range at
every scale, whereas the perplexity-only classifier is substantially weaker. In
ProGen2, this gap widens with scale: the embedding probe improves, while the
perplexity-only classifier drops to AUC $0.72$ at 6.4B. EvoDiff perplexity denotes
the OA-ARDM ELBO-based estimate.}
\label{fig:crossarch_probe}
\end{figure*}

We next repeat the probe-versus-perplexity comparison of Section 4.3.
outside the ESM family (Figure~\ref{fig:crossarch_probe}). For each scale of ProGen2
and EvoDiff, we train a logistic-regression probe on mean-pooled embeddings and compare
it with a perplexity-only zero-shot classifier, $s=-\mathrm{PPL}$, and the same
shallow-feature baselines. All scores are computed on the same
held-out human viral/cellular test split.

The embedding probe remains near ceiling across both non-ESM architectures and improves
with scale. ProGen2 increases from AUC-ROC $0.975$ at 151M parameters to $0.993$ at
6.4B, while EvoDiff increases from $0.965$ at 38M to $0.986$ at 640M. These values fall
within the range of the 13 ESM embedding probes ($0.967$--$0.997$) and remain well above
the best shallow-feature baseline (AUC $0.883$, dipeptide composition). In contrast,
the perplexity-only classifier is substantially weaker. For ProGen2, its AUC decreases
from $0.82$ at 151M to $0.72$ at 6.4B, despite the embedding probe improving over the
same scale range. For EvoDiff, the PPL classifier improves from $0.78$ to $0.88$, but
remains at or below the shallow-feature baseline. Thus, as in the ESM family, scaling
can make the embedding representation more linearly informative while leaving
perplexity-based discrimination weak or even worse at larger scales. 

The signal generalizes to held-out viral families. Leave-one-family-out
evaluation over the 13 viral families with at least 50 sequences gives mean held-out
AUC $0.976$ for ProGen2 and $0.970$ for EvoDiff, close to the full-probe performance. This mirrors the family-level robustness analysis in
Appendix~\ref{app:robustness}: the probe is not simply memorizing the dominant viral
families, but recovers a viral signal that transfers across family boundaries. Together,
these results show that residual viral separability beyond reconstruction difficulty is
not specific to the masked-LM ESM family.

\medskip
\noindent\textbf{Summary.}
Across one autoregressive and one diffusion pLMs, the nativeness axis and residual viral
embedding signal both reproduce outside the ESM family. Scaling also remains
heterogeneous across viral families, but the family ranking changes across objectives.
These results suggest that the main findings are not artifacts of masked language
modeling.

\end{document}